\documentclass[Afour,sageh,times]{sagej}

\usepackage{moreverb,url}
\usepackage{array}
\usepackage{textcomp}
\usepackage{stfloats}
\usepackage{url}
\usepackage{verbatim}
\usepackage{graphicx,wrapfig,lipsum}
\usepackage{amsmath}
\usepackage{tabularx} 
\usepackage{amsthm}
\usepackage{bm}
\usepackage{booktabs}
\usepackage{subcaption}
\usepackage{multirow, booktabs}

\usepackage{algorithm}% http://ctan.org/pkg/algorithm
\usepackage{algpseudocode}
\usepackage{amssymb}
\usepackage{xcolor}

\usepackage{amsthm}
\theoremstyle{definition}
\newtheorem{definition}{Definition}

\usepackage{url}
\usepackage{breakurl}

\usepackage[colorlinks,bookmarksopen,bookmarksnumbered,citecolor=red,urlcolor=red]{hyperref}

\newcommand\BibTeX{{\rmfamily B\kern-.05em \textsc{i\kern-.025em b}\kern-.08em
T\kern-.1667em\lower.7ex\hbox{E}\kern-.125emX}}

\def\volumeyear{2025}
\begin{document}

\runninghead{Dong et al.}
% use `\textit{et~al.}' if there are three or more authors.

\title{Robustness of Robotic Manipulation: \\Foundations and Frontiers
}

\author{Yifei Dong\affilnum{1,2}, Zhanyi Sun\affilnum{3}, Lujie Yang\affilnum{4}, Manuel Baum\affilnum{5}, Kei Ikemura\affilnum{2}, Shuran Song\affilnum{3}, Florian T. Pokorny\affilnum{2}, and Xianyi Cheng\affilnum{1}
}

\affiliation{
\affilnum{1}Duke University, USA\\
\affilnum{2}KTH Royal Institute of Technology, Stockholm, Sweden\\
\affilnum{3}Stanford University, USA\\
\affilnum{4}Massachusetts Institute of Technology, USA\\
\affilnum{5}Organifarms, Germany
}

\corrauth{Yifei Dong, KTH Royal Institute of Technology, Stockholm, Sweden.}

\email{yifeid@kth.se}

\begin{abstract}
Humans and animals exhibit remarkable robustness in physical manipulation, yet robots remain far behind. Progress toward human-level manipulation robustness is hindered by the absence of a unified and systematic understanding: different subfields frame robustness in distinct ways, often leaving the concept ambiguous and limiting deeper analysis as well as communication across research areas. This paper presents a systematic study of manipulation robustness. We begin with a formal definition, characterizing robustness as the degree to which a manipulation system can achieve its goal in the presence of uncertainty and variation. Building on this definition, we introduce general formulations of manipulation robustness from probabilistic and control-theoretic perspectives. We then synthesize the guiding principles and concrete mechanisms of manipulation robustness across perception, planning, control, policy learning, and hardware, illustrating each mechanism through representative works, including foundational and recent studies. In addition, we revisit existing metrics and evaluation methods for quantifying manipulation robustness. Finally, we distill broader lessons for designing robust manipulation systems and discuss open problems and future directions toward achieving human-level robustness in robotic manipulation.
\end{abstract}


\maketitle

\section{Introduction}
Manipulation in the real world is inherently uncertain~\citep{mason2018toward}. Object pose may be partially observed, friction and compliance may be poorly modeled, contact events are discontinuous and difficult to predict, and task instances may vary substantially across episodes. Robust manipulation is therefore not simply a matter of executing a nominal plan accurately in an idealized environment. Instead, it is the ability to \emph{achieve task goals despite such uncertainty and variation}.

\begin{figure*}
\centering
\includegraphics[width=0.999\linewidth]{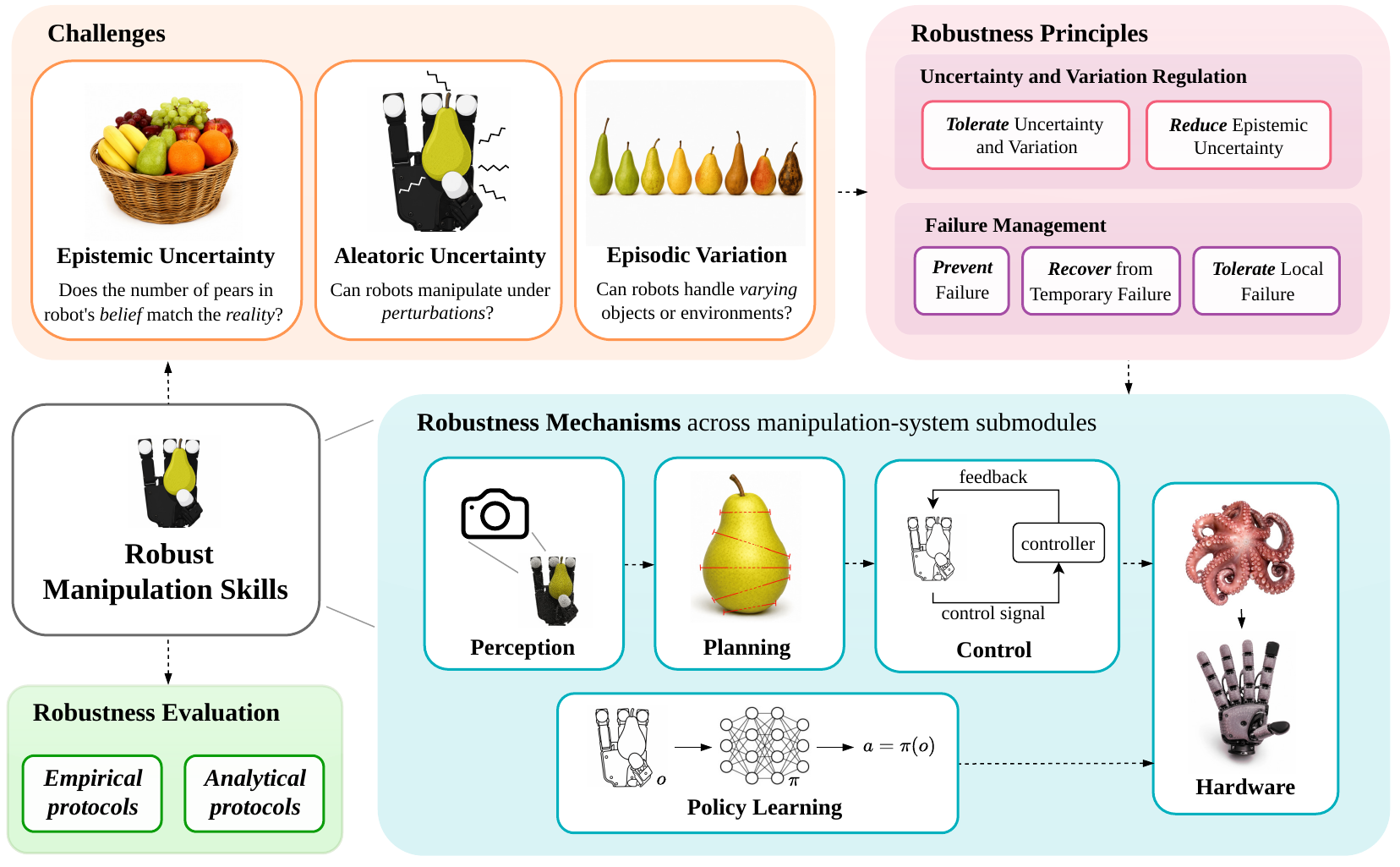}
\caption{
Overview of the robustness of robotic manipulation, illustrated using a pear grasping task. Robotic manipulation faces three core challenges: epistemic uncertainty, aleatoric uncertainty, and episodic variation. The paper is organized around two robustness principles: \emph{uncertainty and variation regulation}, which concerns reducing epistemic uncertainty or tolerating uncertainty and variation, and \emph{failure management}, which concerns preventing failure, recovering from temporary failure, or tolerating local failure. We discuss robustness mechanisms across five submodules of a manipulation system: perception, planning, control, policy learning, and hardware. Several icons or illustrations are original and conceptually inspired by \citep{lu2025grasping,fruithunters_banana_variety_box,onlinedelivery_mixed_fruit_basket_2kg,bair_dex_manip_image1,cao2022real,sankar2025natural}.
% Photo credits: Fruit manipulation with Allegro and human hand~\citep{lu2025grasping} (\copyright\ IEEE); 
% octopus-inspired compliant hand~\citep{sankar2025natural} (\copyright\ AAAS); 
% bananas of varying shapes~\citep{fruithunters_banana_variety_box}; 
% basket of fruits~\citep{onlinedelivery_mixed_fruit_basket_2kg};
% dexterous apple grasping~\citep{bair_dex_manip_image1}; 
% banana grasp synthesis~\citep{cao2022real} (\copyright\ Elsevier).
% (Permission to adapt some photos used in this paper is being requested from the original authors.)
% \yifei{we can add evaluation protocols.}
}
\label{fig:overview}
\end{figure*}

Biological systems provide compelling evidence that robust manipulation is possible. Humans often handle these uncertainties and variations without explicit deliberation, relying instead on sensorimotor control strategies that combine prediction with rapid feedback correction~\citep{flanagan2006control}.
For instance, humans regulate grip forces to prevent objects from slipping in the hand. When grasping objects such as fruits (Fig.~\ref{fig:overview}), humans apply grip forces with a safety margin above the expected slip threshold. Under varying or dynamic conditions, this margin is adjusted accordingly~\citep{hadjiosif2015flexible}, preventing grasp failure even in uncertain environments.

Robust behaviors are also generally observed in animals. Sea otters, for example, juggle pebbles on their bellies~\citep{foster1982captive}. They maintain this behavior despite turbulent disturbances and across wide variations in object shape, mass, and body posture. 
% Rats adjust the speed and pattern of whisking with their vibrissae to increase information about the shape and texture of contacted objects (Fig.~\ref{fig:front}-b), demonstrating how exploratory behavior reduces uncertainty about the environment~\citep{grant2009active}.
Even far less complex organisms exhibit robust interaction with their environment. For example, coordinated ciliary motion in {paramecium} generates fluid flows that draw and capture food particles~\citep{fenchel1980suspension}. Without complex visual perception, the organism can handle uncertainty in particle location using only mechanosensory cilia in highly dynamic fluid environments. These examples suggest that robustness in physical interaction is a fundamental property of biological systems~\citep{kitano2004biological}.

% \begin{figure}
% \centering
% \includegraphics[width=0.99\linewidth]{fig/survey-front.pdf}
% \caption{
% % Examples of robust physical interaction observed in animals. 
% From robust physical interactions observed in animals to robust robotic manipulation skills.
% (a) Sea otters juggle stones on their bellies, maintaining reliable control despite disturbances, or variations in object shape and mass (figure inspired by~\citep{cogdon2020otter_juggling}).
% (b) Robots have demonstrated a degree of robustness, as illustrated by the task of grasping a pear from a cluttered fruit basket. However, significant challenges remain before robotic systems can achieve the level of robustness routinely exhibited by animals and humans.
% % (b) Rats actively explore their surroundings by whisking objects with their vibrissae, gathering information to reduce uncertainty about object shape and location~\citep{zweifel2021dynamical} (\copyright\ PNAS). 
% % (b) {Paramecium} generates ciliary flows that draw and capture food particles despite uncertainty in particle location~\citep{mackean_paramecium_feeding} (\copyright\ D.\,G.~Mackean; annotated with permission).
% }
% % , Ian Mackean
% \label{fig:front}
% \end{figure}

In contrast, today's robotic systems manipulate objects with a level of robustness still below that of humans and many animals~\citep{mason2018toward,billard2019trends}. Even routine behaviors such as pick-and-place, which humans perform effortlessly, remain challenging for robots to execute reliably in the presence of contact uncertainty~\citep{rodriguez2021unstable}. Although impressive demonstrations from academia and industry have shown remarkable capabilities under controlled conditions, many of these systems degrade significantly when deployed outside carefully engineered environments. Ultimately, robotic manipulation must operate in open, time-varying environments characterized by significant uncertainty.

This evident gap between the robustness of robotic and human manipulation motivates this paper. A major obstacle to progress is the absence of a deep understanding and overarching framework for manipulation robustness. Each subfield of robotics discusses robustness in its own terms, often restricted to a narrow domain and leaving the concept implicit or underspecified~\citep{moos2022robust,skogestad2005multivariable,ghazi2017morphological}. To address this issue, we propose a shared conceptual language and a taxonomy to connect existing research threads and enable the accumulation and accessibility of insights across subfields. 
We also identify several foundational but unanswered questions: What explains the extraordinary robustness observed in human and animal manipulation? What mechanisms and principles enable robust manipulation? How should robustness be quantified and measured? This paper calls attention to these questions and encourages the field to pursue a deeper and more systematic understanding of robust manipulation skills.

Robust manipulation skills appear not to arise from a single mechanism, but rather from the coordination of mechanisms across perception, planning, control, policy learning, and hardware, calling for joint research across subfields~\citep{billard2019trends}. This perspective distinguishes the present paper from prior reviews on robotic manipulation, such as data-driven grasp synthesis~\citep{bohg2013datas}, robot learning for manipulation~\citep{kroemer2021review}, foundation models for robotics~\citep{firoozi2025foundation}, etc. While these works have made significant contributions to broad areas of manipulation, robustness is typically treated only incidentally rather than as the central object of study. To our knowledge, there has not yet been a comprehensive review that systematically examines the principles, mechanisms, and evaluation methods underlying robust manipulation under real-world uncertainty and variation.

This paper makes the following \emph{contributions}: (i) it proposes a task-centered definition of manipulation robustness; (ii) it formulates manipulation robustness from both probabilistic and control-theoretic perspectives; (iii) it organizes robustness mechanisms across perception, planning, control, policy learning, and hardware under a set of guiding principles; and (iv) it synthesizes evaluation protocols and open challenges for future research.

In the following, we begin by introducing a definition (Section~\ref{sec:definition}) and a mathematical formulation (Section~\ref{sec:formulation}) of manipulation robustness. We then present robustness principles (Section~\ref{sec:principle}), followed by a systematic summary of existing robustness mechanisms (Section~\ref{sec:survey}). Next, we analyze current metrics and evaluation protocols (Section~\ref{sec:evaluation}). We then synthesize key insights and observations (Section~\ref{sec:discussion}). Finally, we highlight open challenges and suggest future directions toward achieving human-level manipulation robustness (Section~\ref{sec:challenge}).

\section{Definition of Manipulation Robustness}
\label{sec:definition}
% \yifei{This is becoming more of a position paper than a survey.}
This section introduces the concept of robustness progressively. We begin with general definitions of robustness across disciplines, narrow the scope to robotics, and finally formalize the notion of manipulation robustness.

\subsection{Robustness Definitions}
Despite its frequent use, the notion of robustness is often interpreted differently across contexts and disciplines. Several domain-specific efforts have attempted to formalize robustness in biology~\citep{kitano2004biological}, machine learning~\citep{braiek2025machine}, and robotics~\citep{baum2024robustness}. We provide representative definitions below for illustration. 

\begin{definition}[Biological robustness]
Robustness is a property that allows a system to maintain its functions despite external and internal perturbations~\citep{kitano2004biological}.
\end{definition}
\begin{definition}[Machine learning robustness]
The robustness of machine learning models denotes the capacity of a model to sustain stable predictive performance in the face of variations and changes in the input data~\citep{braiek2025machine}.
\end{definition}
\begin{definition}[Robotic robustness]
Robustness specifies the degree to which a behavior can fulfill its task despite
being challenged by an adversity~\citep{baum2024robustness}.
\end{definition}

Synthesizing these perspectives reveals a common theme: robustness is inherently \emph{context-dependent}. It is not an absolute property of a system, but a relational one. It is defined only with respect to what the system is intended to achieve and the conditions under which it operates. Without context, claims of robustness become ill-posed and difficult to compare across domains or methods.

To make this dependence explicit, we identify four foundational dimensions that together define the context of robustness: \textit{goal}, \textit{challenge}, \textit{mechanism}, and \textit{evaluation}. With these dimensions, we define robustness as follows:
\begin{definition}[Robustness]
Robustness refers to a system’s ability to achieve a given \textit{goal} under specified \textit{challenges}, enabled by particular \textit{mechanisms}, and assessed through quantitative \textit{evaluation}.
\label{def:robustness}
\end{definition}

This structure applies broadly across disciplines. For example, in machine learning, robustness typically refers to maintaining predictive performance (\textit{goal})~\citep{braiek2025machine} under distribution shifts or adversarial perturbations (\textit{challenge})~\citep{madry2018towards}, achieved through techniques such as adversarial training (\textit{mechanism}), and evaluated using quantitative metrics such as worst-case accuracy, or performance degradation under controlled perturbations (\textit{evaluation})~\citep{hendrycks2019benchmarking}.

%Across disciplines, the notion of robustness is often discussed with vagueness, despite domain-specific attempts to define it in biology~\citep{kitano2004biological}, machine learning~\citep{braiek2025machine}, and robotics~\citep{baum2024robustness}. This lack of precision leads to unclear statements, overlaps with related concepts, and limits cross-field insight.
%In robotics and related areas, efforts have sought clearer definitions. Baum~\citep{baum2024robustness} surveys robustness across domains and describes it as the ability of a system to maintain performance toward a goal despite perturbations. For robotic behavior, robustness is defined as the degree to which a task can be fulfilled under adversity. Such definitions emphasize context-dependence: robustness must be stated relative to a goal, the challenges considered, and the mechanisms enabling it. A less emphasized but essential point is that robustness is measurable and should be evaluated quantitatively.
% From these perspectives, four foundational dimensions of robustness emerge: goal, challenge, mechanism, and evaluation. This structure applies broadly across disciplines. For instance, in machine learning, robustness refers to predictive performance (goal) under distribution shifts or adversarial perturbations (challenge), achieved through techniques such as adversarial training (mechanism), and assessed by performance under challenging conditions (evaluation).
% \xianyi{ I rewrite this section as follows. Yifei, please add to those examples definitions, and feel free to improve the format and writing.}

\subsection{Robustness in Robotics}
What distinguishes robotics is the embodied physical interactions and tight integration of subsystems. 
\textit{Goals}: Because robotic systems integrate sensing, mechanics, planning, control, actuation, and learning, robustness must often be addressed both within and across the components. 
Thus, goals may be defined at different levels of abstraction, ranging from high-level task completion (e.g., object transport or assembly) to component-level objectives such as state estimation accuracy, collision avoidance, or trajectory tracking performance. 
\textit{Challenges}: Direct interaction with the physical world exposes robots to uncertainty in perception and contact dynamics, sensing noise and external disturbances, as well as variations in system parameters and environmental conditions.
\textit{Mechanisms}: Robustness is realized through complementary passive and active strategies spanning hardware and algorithms. Passive strategies, such as compliant hardware and soft materials, physically tolerate disturbances. Active strategies, such as motion planning, feedback control, and learning methods, proactively mitigate model mismatch or compensate for noise or disturbances.
\textit{Evaluation}: Robustness in robotics is commonly evaluated on task execution success rates, complemented by more fine-grained metrics such as stability measure, control convergence, or theoretical guarantees, defined at the level of individual tasks or system components.

\subsection{Robotic Manipulation Robustness}
Manipulation is characterized as ``an agent’s control of its environment through selective contact''~\citep{mason2018toward}, which makes manipulation robustness a distinct subclass of robotic robustness, with unique challenges arising from contact interactions among the robot, manipulated objects, and the environment.

Manipulation is dominated by multi-body contact interactions, which induce dynamics that are high-dimensional, hybrid, discontinuous, constrained, and therefore difficult to model accurately. These interaction dynamics constitute a major \textit{challenge}, as small variations in geometry, contact conditions, or material properties can lead to qualitatively different outcomes. Perception introduces additional challenges in manipulation because effective contact interaction depends on estimating interaction-relevant states that are largely external to the robot, such as object pose, insertion depth, local contact geometry, material properties, and deformation. Many of these quantities are only partially observable, visually ambiguous, or occluded during contact. Therefore, it has substantially higher demands on perception than tasks such as locomotion, where the robot primarily regulates its own state. Together, these factors complicate the design of robust \textit{mechanisms} in manipulation planning, control, and learning. 

Moreover, real-world manipulation encompasses a broad spectrum of task specifications, in which robots must achieve diverse objectives across domestic, industrial, and open-world environments. Unlike locomotion, manipulation relies on multi-body contact interactions among the robot, manipulated objects, and the environment that actively change the state of the world, leading to a combinatorial diversity of manipulation \textit{goals}. Finally, the discontinuous and task-dependent nature of contact interactions makes the definition of general \textit{evaluation} criteria and the establishment of formal robustness guarantees particularly difficult.

We refine Definition~\ref{def:robustness} for manipulation by specifying the four foundational dimensions as follows:
\begin{itemize}
    \item \textit{Goals}: Robotic manipulation robustness is defined relative to manipulation goals, which involve achieving desired object states through contact interactions, potentially within task-specific tolerance levels.
    \item \textit{Challenges}: Challenges arise from uncertainty and variation, which we categorize into epistemic uncertainty, aleatoric uncertainty, and episodic variation. 
    \item \textit{Mechanisms}: Robustness mechanisms improve task goal achievement by tolerating or mitigating uncertainty and variation in contact interactions, preventing failures, or enabling recovery when failures occur.
    \item \textit{Evaluation}: Evaluation methods define quantitative criteria for measuring manipulation performance and robustness, enabling principled comparison across approaches.
\end{itemize}

Among these dimensions, challenges play a central role. The importance of uncertainty has been recognized since the earliest days of robotics in the 1950s~\citep{goertz1952fundamentals}, which remains a central theme in robotics research~\citep{mason2012creation}. Specifically,
epistemic uncertainty stems from incomplete or inaccurate knowledge of system properties and can, in principle, be reduced through additional data or modeling. Aleatoric uncertainty reflects irreducible randomness, such as observation noise or stochastic transition disturbances. Episodic variation refers to systematic differences in the physical world that remain fixed within a single task execution but change across task episodes, such as variations in object identity, environment configuration, or robot embodiment~\citep{cui2021toward}. An intuitive analogy is that of a robot searching for a light switch in a dark room. Episodic variation: across different rooms, the position or shape of the switch may vary. Epistemic uncertainty: within a given room, the initial belief about the switch location is uncertain but can be refined through interaction, such as touching the wall. Aleatoric uncertainty: even with perfect knowledge, however, randomness such as sensor noise, slips of the hand, or minor actuation errors may still occur.

\section{Formulation of Manipulation Robustness}
\label{sec:formulation}

Given the definition in Section \ref{sec:definition}, we formulate the general problem of manipulation robustness using a unified partially observable stochastic control model. We then instantiate this model through probabilistic and control-theoretic perspectives, respectively. 
The unified model serves as an analytical framework to structure the discussion in the remainder of the paper. It is intended as a conceptual tool for analysis rather than a prescriptive model; it does not imply that a robot must explicitly represent these processes to achieve robust manipulation.

\subsection{Unified Formulation}
We formalize a manipulation episode as a discrete-time control process over a finite horizon $T$, denoted by the sequence $\{(x_t, u_t, y_t)\}_{t=0}^T$. Here, $x_t \in \mathcal{X}$ is the system state (encompassing both robot and object generalized coordinates), $u_t \in \mathcal{U}$ is the control input, and $y_t \in \mathcal{Y}$ is the partial observation.
Note that we present here a time-discretized model, though in principle a continuous-time formulation can also be adopted.

The evolution of this system is driven by underlying deterministic functions subjected to noise:
\begin{align} \label{eq:unified_dynamics}
x_{t+1} &= f(x_t, u_t, w_t ; \theta), \\
y_t &= g(x_t, v_t ; \theta).
\end{align}
In this framework, manipulation robustness is defined by how well a system achieves a target goal~$\mathcal{G}$ despite three fundamental challenges:

\paragraph*{\textbf{Aleatoric Uncertainty}} The variables $w_t \in \mathcal{W}$ and ${v_t \in \mathcal{V}}$ represent exogenous process and measurement noise. This captures inherent, irreducible stochasticity in the physical world, such as thermal sensor noise, unmodeled micro-impacts, or random slip during contact.

\paragraph*{\textbf{Epistemic Uncertainty}} The parameter $\theta \in \Theta$ dictates the true physical and perceptual properties of the task, such as object mass, surface friction, and camera calibration. However, the robot rarely has perfect knowledge of $\theta$ or the true state $x_t$. It operates using internal estimates $\hat{\theta}$ and $\hat{x}_t$. The discrepancy between reality ($\theta, x_t$) and the robot's belief ($\hat{\theta}, \hat{x}_t$) constitutes epistemic uncertainty. Unlike aleatoric noise, this uncertainty can theoretically be reduced through exploration or learning.

\paragraph*{\textbf{Episodic Variation}} While epistemic uncertainty concerns what the robot does not know, episodic variation describes the objective physical differences between separate executions of a task. We define an episodic variation $\nu$ as a specific draw of the environment parameters, initial state, and goal region:
\begin{equation}
\nu = (\theta, x_0, \mathcal{G}) \in \mathcal{N},
\label{eq:variation-def}
\end{equation}
where $\mathcal{N}$ denotes the space of all possible task episodes.
% $\mathcal{N} := \Theta \times \mathcal{X} \times \mathcal{G}$
% , with $\Theta$ representing the set of environment and system parameters, $\mathcal{X}_0$ the set of admissible initial states, and $\mathcal{G}$ the family of feasible goal regions.
% Note that the goal $\mathcal{G} \subset \mathcal{X}$ represents the desired state of, in particular, the manipulated object, consistent with the definition of manipulation as a robot’s interaction with the environment to change a target object's state. Importantly, this notation encompasses a broad range of objectives, and may include trajectory-level subgoals, for example in robotic handwriting or drawing.
Note that the goal $\mathcal{G} \subset \mathcal{X}$ represents a desired set of states, especially the desired state of the manipulated object. This compact notation covers many terminal manipulation objectives, but it does not by itself encode temporal ordering among subgoals. Ordered long-horizon tasks, such as opening a door, entering a room, and then closing the door, would require a richer formulation, e.g., a sequence of goal sets $(\mathcal{G}_1,\ldots,\mathcal{G}_K)$ or a temporal task specification.
% deformable object shape control or 

%%%%%%%%%%%%%%%%%%%%%%%%
\begin{table*}[ht]
\centering
\caption{Overview of the manipulation robustness mechanisms (Learning mechanisms in Table~\ref{tab:learning_robustness_mechanisms}). 
}
% \footnotesize
\label{tab:robustness_mechanisms}
\begin{tabular}{p{1.8cm}p{5.8cm}p{5.5cm}p{1.1cm}}
\toprule
\textbf{Submodule} & \textbf{Mechanism} & \textbf{Principle} & \textbf{Section} \\
\midrule

\multirow{3}{*}{Perception}
& Active and Interactive Perception & Uncertainty Reduction & \ref{sec:perception:interactive} \\ % reduction
& Invariance in Perceptual Representations & Uncertainty \& Variation Toleration & \ref{sec:perception:invariance} \\ % toleration
& Multimodality & Local Failure Toleration & \ref{sec:perception:multimodality} \\ % local failure

\midrule

\multirow{4}{*}{Planning}
& Reasoning over Uncertainties & Uncertainty Reduction & \ref{subsec:planning-uncertainty} \\
& Motion Strategy and Funneling & Uncertainty Reduction & \ref{subsec:planning-funnel} \\
& Robustness Margin & Uncertainty \& Variation Toleration & \ref{subsec:planning-manifold} \\
& Closure & Uncertainty \& Variation Toleration & \ref{subsec:planning-closure} \\
\midrule

\multirow{3}{*}{Control}
% & Reactivity & AU/EV & \ref{control:reactivity} \\
& Predictivity & Failure Prevention & \ref{control:predictivity} \\
& Active Compliance & Uncertainty \& Variation Toleration & \ref{control:compliance} \\
& Control Invariance & Uncertainty \& Variation Toleration & \ref{control:invariance} \\
\midrule

\multirow{3}{*}{Hardware}
& Passive Compliance & Uncertainty \& Variation Toleration & \ref{mechanical:passive-compliance} \\
& Adhesion & Uncertainty \& Variation Toleration & \ref{mechanical:adhesion} \\
% & Postural Synergy & AU/EV & \ref{mechanical:synergy} \\
& Morphology Adaptation & Temporary Failure Recovery & \ref{mechanical:morphology-adaptation} \\
\bottomrule

\end{tabular}
\end{table*}
%%%%%%%%%%%%%%%%%%%%%%%%%%%%%%%%%%

% & Bayesian Perception and Robust Estimation & Uncertainty reduction & \ref{sec:perception:bayesian} \\ % reduction
% Recovery from temporary failure / Failure prevention
% Uncertainty reduction / U/V toleration

% \multirow{4}{*}{Learning}
% & Distribution-centric Robustness &  & \ref{learning-distribution-centric} \\
% & Architecture- and Representation-centric Robustness &  & \ref{architecture-distribution-centric} \\
% & Objective-centric Robustness &  & \ref{objective-distribution-centric} \\
% & Adaptation-centric Robustness & Temporary failure regulation & \ref{adaptation-distribution-centric} \\
% \midrule

\subsection{Probabilistic View}
Learning-based and probabilistic planning methods treat the noise variables ($w_t, v_t$) and episodic variations ($\nu$) as probability distributions. Consequently, the dynamics and observation models are viewed as stochastic transitions, 
\begin{align}
    x_{t+1} \sim \mathcal{T}(\cdot \mid x_t, u_t; \theta), \ y_t \sim \mathcal{O}(\cdot \mid x_t; \theta).
\end{align}
Robustness in this paradigm is typically formulated as a Partially Observable Markov Decision Process (POMDP) and evaluated as the expected performance of a policy $\pi$ across a distribution of episodes $\rho_\mathcal{N}$:
\begin{equation}
\Gamma(\pi) = \mathbb{E}_{\nu \sim \rho_\mathcal{N}} \bigl[ J_{\nu}(\pi) \bigr],
\end{equation}
where $J_\nu(\pi)$ is the expected return of the trajectory under the true parameters $\theta$, despite the policy acting on its imperfect belief $\hat{\theta}$. 
A policy is defined as a mapping $\pi : \mathcal{B} \to \mathcal{U}$ from beliefs to actions, where a belief $b_t \in \mathcal{B}(\mathcal{X})$ is a probability distribution over the state space $\mathcal{X}$ conditioned on the history of observations and actions.

\subsection{Control-Theoretic View}
In contrast, we follow robust control frameworks (e.g., $H_\infty$ or Robust MPC) and model aleatoric noise $w_t, v_t$ and episodic variations $\nu$ as unknown, deterministic variables confined within bounded sets $\mathcal{W}, \mathcal{V}$, and $\mathcal{N}$. Robustness here is framed as a min-max optimization problem. The goal is to synthesize a policy that guarantees constraint satisfaction and minimizes a cost function $c(x_t, u_t)$ against the worst-case possible realizations of the uncertainties:
\begin{equation}
\min_{\pi} \max_{\nu \in \mathcal{N}, w_t \in \mathcal{W}, v_t \in \mathcal{V}} \sum_{t=0}^{T} c(x_t, u_t).
\label{eq:robust-policy-maxmin}
\end{equation}
While the probabilistic view maximizes \textit{average} reliability across a distribution of environments, the control-theoretic view provides strict performance and safety guarantees in a \textit{worst-case} robustness sense.
% , provided the real-world variations never breach the defined bounds $\mathcal{W}$ and $\mathcal{V}$.

%%%%%%%%%%%%%%%%%%%%%%%%%%%%%%%%%%

\section{Principles of Robust Manipulation}
\label{sec:principle}

% There are diverse mechanisms for robustness, covering perception, planning, control, learning, and hardware.
% At a high level of abstraction, the robustness mechanisms can be broadly grouped along two axes. The first, two strategies for dealing with uncertainty and variation: \emph{reduction} of epistemic uncertainty and \emph{toleration} of uncertainty/variation. 
% The second: failure management, preventative mechanisms that assume failure is not ok; 
% on the other hand, “Failure is ok”, specifically temporary, local failure.

There are diverse mechanisms for achieving robustness in robotic manipulation, spanning perception, planning, control, learning, and hardware. While these mechanisms are often studied separately, many can be understood through several recurring principles.
At a high level, robustness mechanisms can be organized along two axes. The first concerns how systems cope with uncertainty and variation: either by \emph{reducing} epistemic uncertainty or by \emph{tolerating} uncertainty and variation. The second concerns {failure management}: whether robustness is achieved primarily through \emph{preventing} failures from occurring, \emph{recovering} from temporary failures or \emph{tolerating} local failures.

\subsection{Uncertainty and Variation Regulation}

One route to robustness is to \emph{reduce} epistemic uncertainty by improving the robot's knowledge of the environment, task, or system state. This principle underlies active and interactive perception  (Section~\ref{sec:perception:interactive}), belief-space planning methods that explicitly reason over uncertainty (\ref{subsec:planning-uncertainty}), and world-model-based learning approaches that learn predictive representations of environment dynamics (\ref{architecture-distribution-centric}).

In contrast, a second route is based on \emph{tolerating} the challenges: it accepts that aleatoric uncertainty and episodic variation cannot be eliminated, and instead seeks to make task execution insensitive to them. This principle is perhaps more ubiquitously employed in robotic manipulation. One common strategy is \emph{compliance}, realized both actively through control (Section~\ref{control:compliance}) and passively through hardware and materials (\ref{mechanical:passive-compliance}), which reduces sensitivity to contact uncertainty and modeling errors. Another is \emph{invariance}, where behaviors or representations are designed to remain unaffected by uncertainty or variation, as exemplified by closure-based grasping (\ref{subsec:planning-closure}), invariant perceptual representations (\ref{sec:perception:invariance}), and control invariance (\ref{control:invariance}). Robustness can also arise from \emph{mechanics}-driven strategies that exploit task and environmental structure, such as motion strategies and funnels (\ref{subsec:planning-funnel}). Toleration also appears in learning-based methods that generalize across scenes, tasks, and environments through data diversity or policy design (\ref{learning-distribution-centric}, \ref{architecture-distribution-centric}), as well as in hardware mechanisms such as adhesion (\ref{mechanical:adhesion}), which leverage favorable contact physics to tolerate uncertainty and variation.

These two strategies are not mutually exclusive. Robust manipulation systems typically combine them, reducing uncertainty where possible while remaining effective in the presence of irreducible uncertainty.

% adaptation-centric learning that updates policies using deployment data (\ref{adaptation-distribution-centric}). 

% through mechanisms such as active (control) and passive (hardware) compliance (\ref{control:compliance}, \ref{mechanical:passive-compliance}). 
% Motion strategies, margin-based and closure-based strategies in planning (\ref{subsec:planning-funnel}, \ref{subsec:planning-manifold}, \ref{subsec:planning-closure}) can tolerate minor model mismatch or external disturbance by exploit.... 
% invariance - invariant perceptual representations (against sensing variations) (\ref{sec:perception:invariance}) and invariance controllers (against perturbations) (\ref{control:invariance}). 

\subsection{Failure Management}

% Recovery [temporary failure] 
% modularity/redundancy [local failure management]
% adaptation (learning, hardware) [adapt to new conditions], 
% along the other axis, 
% Failure management. Two complementary strategies: preventative strategies that does not allow failure a priori; {failure tolerant} strategies. The relative importance of each depends strongly on task semantics. Tasks involving fragile objects or irreversible consequences often demand predictive failure avoidance, as even brief failure may be unacceptable.  In contrast, in tasks where temporary failure is acceptable, such as regrasping rigid objects after a drop, recovery mechanisms may suffice.
% Failure prevention aims to avoid entering failure-imminent states altogether. including various strategies. Examples include down-weighting risky rollouts in receding-horizon control (Section~\ref{control:predictivity}), synthesizing grasps using closure-based analysis robust to disturbance wrenches (\ref{control:compliance}). exploiting compliance through control and hardware design (\ref{control:compliance}, \ref{mechanical:passive-compliance}), reducing the likelihood of failure due to contact modeling error. Many learned manipulation policies that were trained on successful demonstration trajectories. These approaches emphasize anticipation and constraint satisfaction. 

Along the second axis, robustness can be understood through failure management. Here, two complementary strategies emerge: failure prevention, which seeks to avoid failure a priori, and failure recovery or tolerance, which assumes that failures may occur and emphasizes recovery from them or tolerance of local failure. Tasks involving safety-critical operations or irreversible consequences (e.g., an object's fragility) often demand preventative mechanisms, as failures are unacceptable. In contrast, for tasks where temporary failures can be tolerated, such as regrasping a rigid object after a drop, recovery mechanisms may be sufficient. Local failure tolerance further complements these strategies by allowing failures confined to specific components, sensing modalities, or subtasks to be absorbed without causing overall task failure.

\emph{Failure prevention} aims to avoid entering failure-prone states altogether. This principle appears in a variety of robustness mechanisms. Examples include predictive control methods that down-weight risky rollouts in receding-horizon optimization (Section~\ref{control:predictivity}), closure-based grasp synthesis that remains stable under disturbance wrenches (\ref{subsec:planning-closure}), and compliance-based control and hardware design (\ref{control:compliance}, \ref{mechanical:passive-compliance}), which reduce the likelihood of failure arising from contact modeling errors. Many learning-based manipulation policies similarly emphasize failure avoidance, either by training predominantly on successful demonstrations or by incorporating conservative and safety-aware objectives that discourage risky behaviors and constraint violations (\ref{objective-distribution-centric}).
% Across these approaches, robustness is achieved through anticipation, conservatism, and constraint satisfaction.

% By contrast, failure recovery focuses on restoring task execution once failure has occurred. failure can be temporary and then recovered. recent learning-based systems demonstrate the ability to acquire recovery behaviors from imperfect or failed demonstrations (Section \ref{learning-distribution-centric}), or specific recovery policies from failure. 
% temporary failure might be due to shift of distribution or condition, therefore adaptation needed, no matter morphological adaptation (\ref{mechanical:morphology-adaptation}) or learning-based adaptation (\ref{adaptation-distribution-centric}).
% On the other hand, failure can be local, and we can alleviate local failure through modularized or redundant system design. examples are modularized robot~\citep{yim2000polybot,xie2019reconfigurable}, localize disturbances and prevent their propagation, enhance fault tolerance. 
% \emph{redundancy}, provides multiple pathways (often more than needed) to task success and allows local degradation in one component to be compensated by others.
% redundancy examples - multi-sensor fusion (\ref{sec:perception:multimodality}), dual end-effectors~\citep{zeng2022robotic}, and alternative task-level plans~\citep{simeon2004manipulation}.
% modularity and redundancy are often bio-inspired, like ants dynamically reassign roles during cooperative transport with redundant individuals for a given subtask~\citep{gelblum2015ant}.

By contrast, a complementary strategy focuses on restoring task execution once failure has occurred. From this perspective, failures need not be catastrophic; they may be temporary and recoverable, or local and confined to particular components. Recent learning-based systems have demonstrated the ability to acquire recovery behaviors from imperfect or failed demonstrations (Section~\ref{learning-distribution-centric}), or to explicitly learn recovery policies that resume task execution following failure. \emph{Temporary failures} may also arise from distribution shifts or changing task conditions, motivating adaptation mechanisms, either through morphological adaptation (\ref{mechanical:morphology-adaptation}) or learning-based adaptation (\ref{adaptation-distribution-centric}). 

\emph{Local system failures}, on the other hand, refer to failures confined to a particular component, sensing modality, actuator, module, or subtask, without immediately implying failure of the overall manipulation task. Such failures can be mitigated through modularity and redundancy. Modular robotic systems~\citep{yim2000polybot,xie2019reconfigurable} can localize failure and prevent its propagation to the rest of the system. Redundancy provides multiple pathways to task success and allows degradation in one component to be compensated for by others. Examples include multi-sensor fusion (Section~\ref{sec:perception:multimodality}), dual end-effectors~\citep{zeng2022robotic}, and alternative task-level plans~\citep{simeon2004manipulation}. Both modularity and redundancy are also common principles in biological systems; for example, ants dynamically reassign roles during cooperative transport, using redundant individuals to maintain collective performance despite local failures~\citep{gelblum2015ant}.
% Model-based approaches can stabilize payloads after actuator failure~\citep{michael2011cooperative}, while  
% Robust manipulation systems often integrate both strategies, balancing prevention with recovery to handle a wide range of tasks.

Guided by these two axes of robustness principles---uncertainty and variation regulation, and failure management---we next revisit the mechanisms that enable robust manipulation.
\section{Mechanisms of Robust Manipulation}
\label{sec:survey}
In this section, we revisit existing approaches to robustness in robotic manipulation. We organize the literature into five submodules of a robotic system---perception, planning, control, learning, and hardware (Fig.~\ref{fig:overview})---and summarize the mechanisms in Table~\ref{tab:robustness_mechanisms} and Table~\ref{tab:learning_robustness_mechanisms}. Within each submodule, we discuss the mechanisms through the lens of the robustness principles introduced in Section~\ref{sec:principle}, while noting that each mechanism is listed under its \emph{dominant principle} even though it may reflect multiple principles in practice. We also highlight the challenges they address, representative works, and illustrative manipulation tasks. Owing to the rapid recent progress and growing attention to robot learning, the policy learning section (\ref{policy_learning}) covers a comparatively larger body of recent work. At the same time, we emphasize foundational contributions, particularly in planning and control, to modern robotic manipulation.

\subsection{Perception}

In the context of manipulation, the goal of perception is to acquire sufficient information for a robot to interact effectively with objects in its environment. Perception is particularly challenged by epistemic uncertainty, since task-relevant properties and system dynamics are often initially unknown. Important information, such as the number of objects in a scene, their motion constraints, or friction properties, may only become available through interaction with the environment. Consequently, active and interactive perception (Section~\ref{sec:perception:interactive}) constitute important mechanisms for \emph{reducing} uncertainty. At the same time, perception should maintain beliefs only over variables that are relevant to the task while remaining invariant to irrelevant features (\ref{sec:perception:invariance}); such invariance is an effective mechanism for \emph{tolerating} episodic variation across scenes. From a failure-management perspective, multimodal perception provides redundancy across sensing modalities and can mitigate \emph{local failures} within the perception system (\ref{sec:perception:multimodality}).

\begin{figure}
\centering
\includegraphics[width=0.99\linewidth]{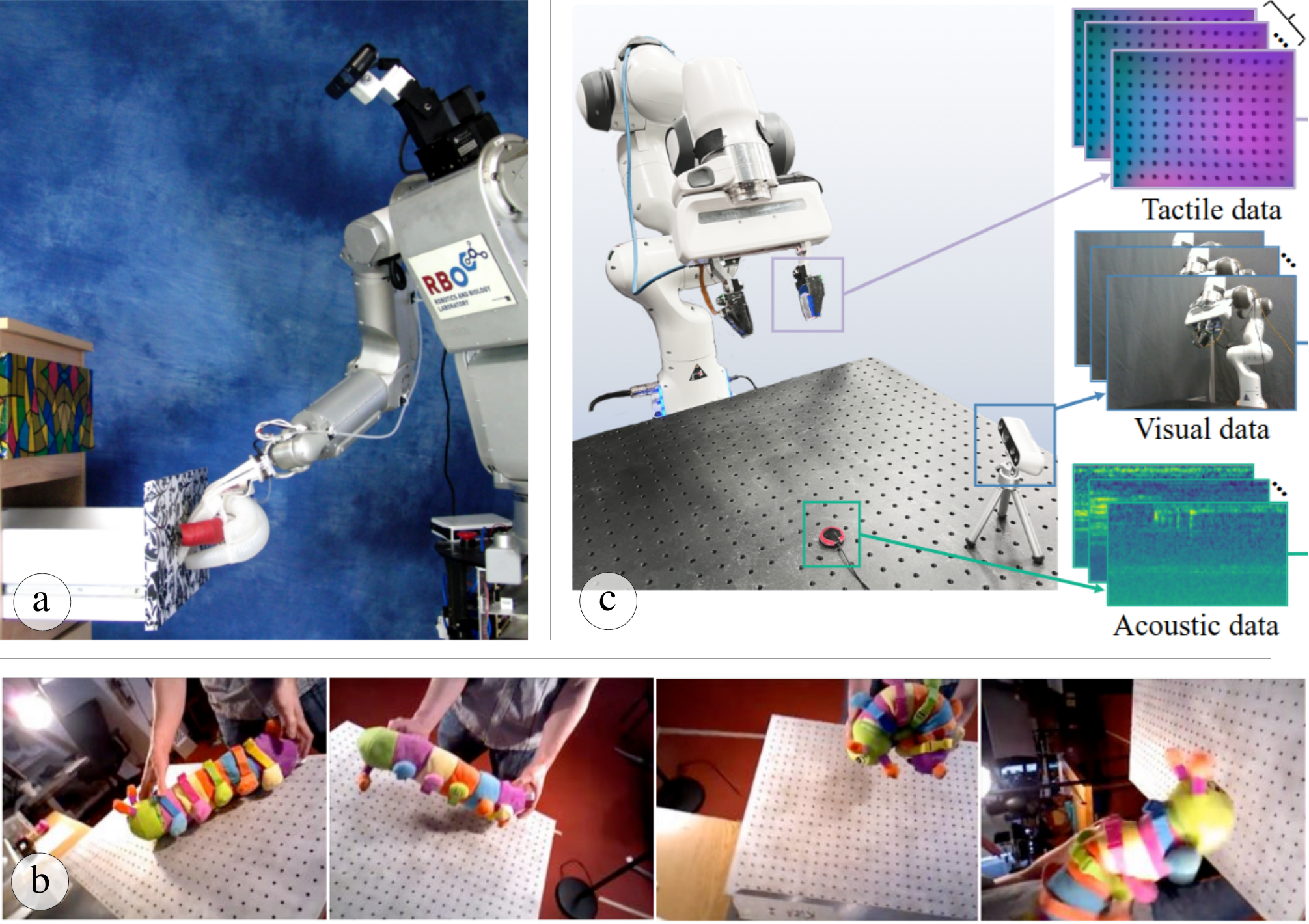}
\caption{Examples of robustness mechanisms in perception. 
(a) Interactive perception: the robot interacts with the environment to improve perceptual understanding~\citep{martin2018leveraging}.
% (c) Bayesian perception and robust estimation: object pose uncertainty is explicitly represented using Bayesian methods~\citep{koval2015pose} (\copyright\ SAGE).
(b) Invariance in perceptual representation: the perception of the toy remains invariant to changes in camera pose, object pose or deformation, and lighting conditions~\citep{florence2018dense}.
(c) Multimodal sensing: perception combining tactile, visual, and acoustic sensing modalities~\citep{li2023see}.
Figures are used with permission from the authors, or adapted under CC BY.
% 3 figs are under 
}
\label{fig:perception}
\end{figure}

\subsubsection{Active and Interactive Perception}
\label{sec:perception:interactive}
Epistemic uncertainty in robot perception can be \emph{reduced} by exploiting a robot's ability to act in and interact with its environment~\citep{aloimonos1988active}. A major challenge is occlusion, which can often be addressed simply by moving a camera to acquire a different viewpoint~\citep{bajcsy1988active}, for example, to support grasp execution~\citep{morrison2019multi} (Fig.~\ref{fig:perception}-a). By actively selecting viewpoints that reveal previously hidden information, perceptual uncertainty about the environment is reduced, facilitating subsequent manipulation.
Beyond contact-free viewpoint selection, robots can also physically interact with their environment to acquire information. This paradigm, known as interactive perception~\citep{bohg2017interactive,xiong2025vision}, is likewise observed in animals that employ exploratory behaviors to gather information about their surroundings~\citep{chappell2012build}. Many task-relevant object properties are difficult or impossible to infer from passive observation alone. For example, lifting a milk carton reveals its mass distribution, while removing the lid of a box may reveal its contents. In cluttered environments, robots may need to manipulate obstructing objects to expose a target object~\citep{xu2015autoscanning}. Interactive perception can also reveal motion constraints and kinematic structures of articulated objects~\citep{katz2008manipulating}. Across these examples, the common principle is the active reduction of epistemic uncertainty through action and interaction.

% Active camera movement can also increase the robustness of structure and depth estimation, even. This works even in highly challenging circumstances, such as seeing objects through a mesh-like fabric~\citep{battaje2022robust}. 
% . In particular, scene segmentation can benefit from pushing actions that induce motion in the scene~\citep{xu2015autoscanning}. 
% However, direct interaction can be costly and risky. A common way to mitigate this is through model-based approaches that predict outcomes of exploratory actions. Such models are often the basis of deliberate exploration strategies, such as entropy minimization or cross-entropy maximization~\citep{baum2017opening}.

\subsubsection{Invariance in Perceptual Representations}
\label{sec:perception:invariance}
Perception for manipulation is challenging because objects need to be reliably detected from different viewing angles and lighting conditions (Fig.~\ref{fig:perception}-b). In this sense, invariant perceptual representations help \emph{tolerate} episodic variations in viewpoint, appearance, and environmental conditions. Before deep learning approaches started to dominate computer vision~\citep{krizhevsky2012imagenet}, computer vision pipelines were often based on features such as SIFT~\citep{lowe2004distinctive}, SURF~\citep{bay2006surf}, or ORB~\citep{rublee2011orb}, which were designed to be invariant to rotation, scale, and possibly other transforms. In deep learning based approaches, such invariance is often achieved using data-augmentation~\citep{krizhevsky2012imagenet} and pooling mechanisms~\citep{lecun2002gradient}. More recently, foundation models such as DINO~\citep{oquab2023dinov2} have demonstrated the ability to learn task-agnostic visual representations that are robust to clutter, occlusion, and illumination changes. At an even higher level of abstraction, language-based scene representations exhibit strong invariance to visual variations, contributing to the robustness and generalization capabilities observed in vision-language-action models~\citep{zitkovich2023rt}.

\subsubsection{Multimodality}
\label{sec:perception:multimodality}
Robotic manipulation can leverage multiple sensing modalities, including visual, tactile, auditory, and proprioceptive information, to perceive and model the environment, thereby enabling more effective interaction with it (Fig.~\ref{fig:perception}-c). These modalities are both complementary and partially redundant in their functions. As a result, failure in one modality need not lead to failure of the overall perception system; other modalities can still provide sufficient information to support task execution. Humans, for example, often rely on touch when vision is unavailable. In robotics, vision provides rich global scene awareness but is susceptible to occlusion, whereas tactile sensing offers detailed local information about properties such as stiffness, texture, friction, and contact geometry. Audition can capture transient interaction events, such as the sound of a snap-fit assembly~\citep{li2023see}. These modalities have been successfully combined in prior work to improve perception robustness~\citep{izatt2017tracking,homberg2019robust}. Beyond multimodal sensing, redundancy can also arise within a single sensing modality, as exemplified by the large number of whiskers in rodents or the compound eyes of insects. From a failure-management perspective, such redundancy helps mitigate \emph{local failures} and prevents them from compromising overall perceptual functionality.

% Multimodality enables robustness because it supports a strong form of \emph{redundancy}.
 % (e.g., linking the instruction ``pick up the mug'' to the correct object)
% For instance, Mirano et al.~\citep{izatt2017tracking} fused \mbox{RGB-D} with tactile feedback for contact-aware pose tracking robust to occlusion. Similarly, proprioception (resistive bend sensors) complements exteroception (fingertip force sensors) in soft hands to improve grasp robustness under uncertainty~\citep{homberg2019robust}. 
% This can be understood as redundancy across modalities: the different sensing channels compensate for one another. 

\subsection{Planning}
% Manipulation planning operates based on perception and is therefore inevitably affected by its imperfections. In particular, imperfect perception or state estimation induces {epistemic uncertainty}: a mismatch between the true robot–environment parameter $\theta$ and the internal model $\hat{\theta}$ (dynamics uncertainty), and consequently between the true state $x_t$ and the state estimate $\hat{x}_t$ (state uncertainty). Under such mismatches, trajectories computed by the planner may fail to achieve the goal $\mathcal{G}$.
% Manipulation planning must also contend with aleatoric uncertainty in the dynamics and observation models, as well as episodic variation across task instances.
% Existing robust planning mechanisms improve manipulation performance through two complementary strategies. On the one hand, epistemic uncertainty can be \emph{reduced} by explicitly reasoning over it (Section~\ref{subsec:planning-uncertainty}) or by exploiting task mechanics through motion strategies and funnels (\ref{subsec:planning-funnel}). 
% Otherwise, planning can \emph{tolerate} uncertainties or disturbances by considering a robustness margin (\ref{subsec:planning-manifold}) or by forming ``closures'' around target objects (\ref{subsec:planning-closure}).

Manipulation planning operates on information provided by perception and is therefore inevitably affected by perceptual imperfections. In particular, imperfect perception and state estimation introduce \emph{epistemic uncertainty}: a mismatch between the true robot--environment parameters~$\theta$ and the internal model $\hat{\theta}$ (dynamics uncertainty), and consequently between the true state $x_t$ and its estimate $\hat{x}_t$ (state uncertainty). Under such mismatches, trajectories computed by the planner may fail to achieve the desired goal $\mathcal{G}$. Manipulation planning must also contend with aleatoric uncertainty in the dynamics and observation models, as well as episodic variation across task instances.

Existing robust planning mechanisms improve manipulation performance through two complementary strategies. On the one hand, epistemic uncertainty can be \emph{reduced} by explicitly reasoning about it (Section~\ref{subsec:planning-uncertainty}) or by exploiting task mechanics through motion strategies and funnels (\ref{subsec:planning-funnel}). On the other hand, planning can \emph{tolerate} uncertainty and disturbances by incorporating robustness margins (\ref{subsec:planning-manifold}) or by establishing forms of \emph{closure} around target objects (\ref{subsec:planning-closure}).

\subsubsection{Reasoning over Uncertainties}
\label{subsec:planning-uncertainty}
Even high-quality sensors and advanced perception algorithms cannot fully eliminate epistemic uncertainty. A direct way to mitigate its effect is to represent uncertainty explicitly as a probability distribution over possible states---commonly referred to as a belief---and to update this belief online using new observations (Fig.~\ref{fig:planning}-a). Formally, the agent maintains a belief $b_t \in \mathcal{B}(\mathcal{X})$ over the state $x_t$, conditioned on the history $h_t = (y_{0:t}, u_{0:t-1})$ under the internal model $\hat{\theta}$, i.e., $b_t(x) = \mathbb{P}(x_t = x \mid h_t, \hat{\theta})$. Planning then operates in belief space through a policy $u_t = \pi(b_t)$, aiming to improve expected task performance~$\Gamma(\pi)$ across episodic variations~$\nu$.

Belief representations enable exploration that actively \emph{reduces} uncertainty by trading off immediate task performance against information gathering. They also support cautious control, allowing robot policies to account for the reliability of available information. For instance, \cite{platt2010belief} applied belief dynamics to grasping, synthesizing locally optimal feedback policies in belief space with replanning. Similarly, \cite{jankowski2024robust} incorporated pose and contact-dynamics uncertainty into a belief-space formulation and validated it on planar pushing without sensory feedback. At longer horizons, belief reasoning extends to uncertainty-aware task and motion planning; \cite{kaelbling2013integrated} integrated belief propagation with symbolic decision making to address incomplete knowledge of object properties and locations in mobile manipulation.
% about object states or system dynamics as a \emph{belief}---

\subsubsection{Motion Strategy and Funneling}
\label{subsec:planning-funnel}
Besides reasoning over uncertainties in belief space, another way to \emph{reduce} uncertainty is to exploit task mechanics. In \emph{motion strategies}, also termed sensorless manipulation by \cite{erdmann2002exploration}, contact dynamics and gravity provide passive negative feedback that drives objects toward desired configurations, often without exteroceptive sensing. This contrasts with sensor strategies, which rely on rich observations but are susceptible to perceptual errors. Motion strategies are effective in diverse settings, including tray-tilting to orient objects of unknown pose using only gravity and contact~\citep{erdmann2002exploration}, push--grasp behaviors in clutter that funnel the target into the gripper while displacing distractors~\citep{dogar2012physics}, and industrial part feeders whose fences enforce consistent orientation under pose and contact variability~\citep{peshkin2002planning}. More generally, environmental contact and gravity as task mechanics can \emph{reduce} uncertainties, or {collapse} episodic variations into a predictable, smaller set of outcomes (often termed \emph{extrinsic dexterity}~\citep{dafle2014extrinsic}). For example, pressing a loose pair of chopsticks against a tabletop passively aligns their tips through the collision.

\begin{figure}
\centering
\includegraphics[width=0.75\linewidth]{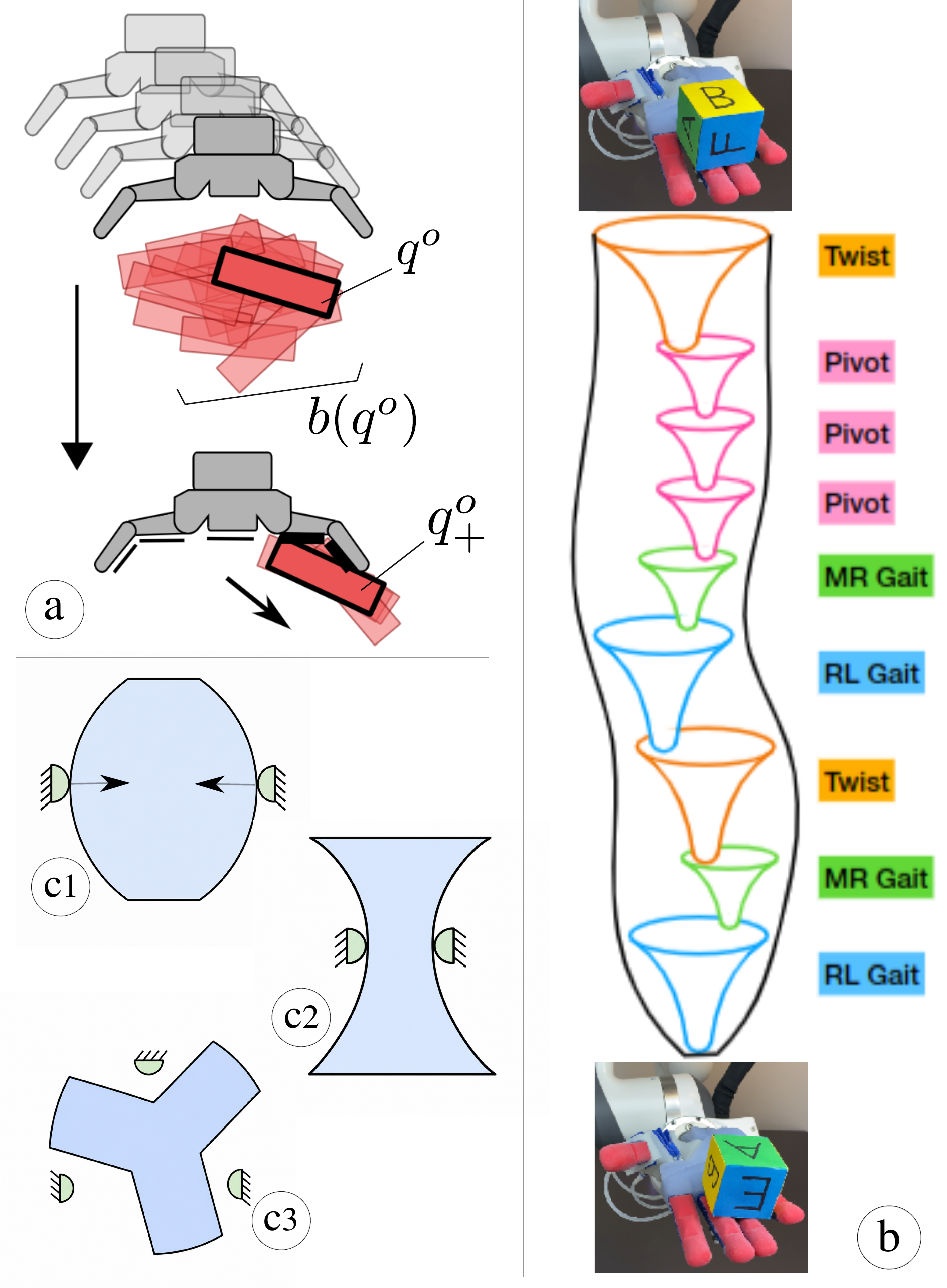}
\caption{Examples of robust planning mechanisms. 
(a) Reasoning over uncertainties: planning in belief space, where the belief of the box position $b(q^o)$ is updated to $b(q^o_+)$ after it is pushed by the robot hand~\citep{koval2016pre}.
% \lujie{Also annotate with b(q) and b(q+) for belief?} 
% (b) Constraint-aware planning: a collision-free trajectory (a–b–c–d–e) is planned on a constrained manifold defined by kinematic constraints (purple) and the constraint of the dumbbell sliding on the table (green)~\citep{berenson2009manipulation} (\copyright\ IEEE).
% \yifei{b can be removed.}
% (b) Motion strategy and funneling: 
% (b1) funneling on a conveyor belt~\citep{peshkin2002planning}, where objects with varying poses converge to a single pose through guiding fences (\copyright\ IEEE); 
(b) Funneling with in-hand reorientation primitives, where primitives are chained to reach the target cube orientation~\citep{bhatt2021surprisingly}; 
% (c3) vibrating bowl feeders, where only parts with a specific orientation can pass through the gate (screenshot from \citep{bowl_feeder_video});
% a \href{https://www.youtube.com/watch?v=8StG7rbspmY}{YouTube video}); 
% (b3) motion strategy for aligning chopsticks~\citep{shi2020hand} (\copyright\ IEEE).
(c) Force closure (c1), form closure (c2), and caging (c3).
Photos in (a) and (b) are adapted under CC BY; figure (c) is inspired by~\cite{fox_grasping_slides,rodriguez2012caging}.  
}
\label{fig:planning}
\end{figure}

A representative abstraction of motion strategy is \emph{funneling}~\citep{mason1985mechanics}. A funnel is a region of attraction $\mathcal{F}\subset\mathcal{X}$ such that, for $x_0\in\mathcal{F}$ and bounded disturbances $w_t \in \mathcal{W}, v_t \in \mathcal{V}$, the resulting execution converges to the goal set $\mathcal{G}$ (i.e., $x_T \in \mathcal{G}$), in other words \emph{reducing} the pose or geometry uncertainty and model mismatch. Such strategies exploit repeated contacts and environmental constraints to steer outcomes without precise sensing, as in vibrating bowl feeders used for industrial part feeding. The funneling concept traces to pre-image backchaining~\citep{lozano1984automatic}, which recursively constructs sets of states from which motions succeed under bounded execution errors. It was later extended by LQR-Tree methods~\citep{tedrake2009lqr}, where verified funnels cover the reachable state space to enable robust trajectory planning under model uncertainty. More recently, composing simple in-hand manipulation funnels has shown surprising robustness to variations in object mass and pose~\citep{bhatt2021surprisingly} (Fig.~\ref{fig:planning}-b).
% Future work may explore hybrid approaches that integrate motion strategies with minimal sensing, combining the robustness of passive mechanics with the adaptability of sparse active feedback for unstructured environments.
% A funnel is a mechanism that mechanically attenuates uncertainty

% \subsubsection{Constraint-aware Planning}
% \subsubsection{Reducing Sensitivity to Uncertainties}
\subsubsection{Robustness Margin}
\label{subsec:planning-manifold}
Beyond explicitly estimating or compensating for uncertainty, an alternative strategy is to \emph{tolerate} uncertainty by planning with a robustness margin. The central idea is to generate motions that remain sufficiently far from constraint boundaries, so that moderate disturbances or model mismatch do not immediately lead to failure. Many manipulation tasks require changing the object's state through interaction while satisfying state and input constraints, $x_t\in\mathcal{X}$ and $u_t\in\mathcal{U}$. These constraints encode collision avoidance, contact consistency, kinematic limits, torque bounds, and related feasibility conditions. Classical methods in this category primarily guarantee feasibility under a nominal model~\citep{berenson2009manipulation,simeon2004manipulation,saha2007manipulation}.

However, under epistemic mismatch in $\theta$ and stochastic disturbances, executions may deviate from the nominal constraint manifold and violate constraints. Robust manipulation planning addresses this gap by constructing robustified feasible subsets, thereby maintaining a ``margin'' from constraint violation to tolerate uncertainty. For example, chance-constrained formulations enforce the probabilistic satisfaction of contact and state constraints under stochastic dynamics~\citep{blackmore2011chance,shirai2022chance}, while set-based approaches employ tightened constraint sets or disturbance-invariant tubes to guarantee feasibility under bounded uncertainty~\citep{mayne2005robust}.
These approaches ensure that constraint satisfaction is preserved within a specified disturbance or risk budget, thereby tolerating uncertainty at the planning level. We will discuss several more works along this line in Section~\ref{sec:eval-margin} and how they can be used as robustness evaluation protocols.
The rich body of work in robust control theory provides a strong foundation for future developments of this mechanism.

\subsubsection{Closure}
\label{subsec:planning-closure}
In the spirit of uncertainty tolerance, robustness can also be achieved through \emph{closure}, which imposes geometric or force constraints on object motion and thereby makes the motion resistant to, for example, external disturbances to manipulated objects. In many grasping and fixturing tasks, the goal is to maintain the object within an in-hand stable set. Therefore, closure can be viewed as shaping the robot-object interactions such that the object’s reachable configurations remain within a bounded subset ${\mathcal{G} \subset \mathcal{X}}$ despite bounded disturbances ${w_t\in\mathcal{W}}$ in the dynamics ${x_{t+1}=f(x_t,u_t,w_t ; \theta)}$. \emph{Force closure} captures equilibrium grasps that can resist arbitrary external wrenches via internal contact forces~\citep{howard1996stability} (Fig.~\ref{fig:planning}-c1). \emph{Form closure} is a geometric condition in which contact constraints eliminate all object motions even without friction~\citep{bicchi1995closure} (Fig.~\ref{fig:planning}-c2). \emph{Caging} (also referred to as ``object closure'' in some contexts) further relaxes contact requirements by trapping the object within a bounded region of its configuration space without precise force regulation~\citep{pereira2004decentralized} (Fig.~\ref{fig:planning}-c3). Across these closure methods, the shared principle is to restrict the configuration space so that the object cannot move freely or escape, thereby making the manipulation outcome \emph{tolerant} to perceptual errors, control imprecision, and external disturbances.

\subsection{Control}
Whereas planning operates at a higher level, often through open-loop decision making, control closes the loop in real time by responding to sensory feedback and correcting deviations that no plan can fully anticipate. A large class of robotic control methods addresses motion control in free space, such as trajectory tracking or point-to-point reaching, where the manipulator operates without contact. Manipulation, however, is fundamentally characterized by the making and breaking of unilateral frictional contacts, which introduce discontinuities and uncertainties. As a result, manipulation control must regulate not only motion but also forces and contact events. Controllers in contact-rich settings must therefore contend with external disturbances and actuation errors, as well as epistemic uncertainty arising from imperfect contact models. Poorly designed controllers can amplify these challenges, leading to instability or excessive contact forces, particularly in interactions with rigid environments. From the perspective of robustness, \emph{compliance} (\ref{control:compliance}) and \emph{invariance} (\ref{control:invariance}) primarily serve to \emph{tolerate} uncertainty and variation, whereas \emph{predictive} control (\ref{control:predictivity}) contributes to \emph{failure prevention} by anticipating future contact events and system evolution.

\subsubsection{Compliance}
\label{control:compliance}
Compliance balances the control of position and force; higher compliance allows larger position deviations under external forces. It can be realized passively through mechanical design (Section~\ref{mechanical:passive-compliance}) or actively within the control loop~\citep{mason2007compliance}. Active compliance control regulates how a robot responds to external forces during contact, allowing it to comply rather than rigidly enforcing a motion. 
This property makes compliance a toleration strategy that bridges the gap between the robot's internal model and the complex, often unpredictable, physics of the real world. Impedance and admittance control represent the two classical approaches. Impedance control enforces compliance from motion to force by commanding motion trajectories that indirectly regulate contact forces~\citep{hogan1984impedance}, 
\begin{equation}
    M_d(\ddot x - \ddot x_d) + D_d (\dot x - \dot x_d) + K_d (x-x_d) = \tau_{\text{ext}},
\label{eq:impedance}
\end{equation}
where $x \in \mathbb{R}^m$ denotes the task-space position (or pose) of the end-effector, $x_d$ is the desired trajectory, and $\tau_{\text{ext}}$ is the external wrench arising from contact. The matrices $M_d$, $D_d$, and $K_d$ are the desired inertia, damping, and stiffness parameters that define a virtual mass–spring–damper system. This equation specifies the closed-loop interaction dynamics: external forces induce bounded motion deviations governed by $(M_d,D_d,K_d)$.
On the other hand, admittance control does so from force to motion by computing the motion response to a commanded force~\citep{dimeas2015reinforcement}. 

Classical active compliance methods typically rely on accurate models or restrictive assumptions. More recently, model-free approaches have emerged, learning compliant manipulation behavior from human demonstrations or through reinforcement learning~\citep{xu2026compliant,hou2025adaptive,kamijo2024learning}. For example, an adaptive compliance policy can adjust compliance both spatially and temporally from demonstrations for contact-rich tasks~\citep{hou2025adaptive}. Future directions may explore combining active compliance with passive mechanical compliance to enhance adaptability under contact uncertainty.

% \lujie{Need more math here. Specifically, an instantiation of $f$.}

\subsubsection{Invariance}
\label{control:invariance}
Invariance ensures that a system executes the behavior prescribed by a nominal controller while remaining within a certified safe set, such as bounds on force tracking error~\citep{polverini2017implicit} or admissible state regions. Like other active compliance control, invariance-based methods do not reduce epistemic uncertainty in states or parameters. Rather, they achieve robustness by constraining system evolution so that bounded disturbances or modeling errors cannot drive the state outside a prescribed safe set; in this way, disturbances are \emph{tolerated} rather than eliminated.
Formally, consider the closed-loop dynamics $x_{t+1} = f(x_t, u_t, w_t; \theta)$ under an output-feedback policy $u_t=\pi_t(b_t)$, where $b_t(x) = \mathbb{P}(x_t = x \mid h_t, \hat{\theta})$. A set $\mathcal{C}\subset\mathcal{X}$ is robustly forward invariant if
\[
x_t \in \mathcal{C} \;\Rightarrow\; x_{t+1}\in \mathcal{C}, \quad \forall w_t\in\mathcal{W}.
\]
Invariance-based control enforces this condition by modifying or projecting control inputs when the boundary of $\mathcal{C}$ is at risk of being violated, thereby guaranteeing constraint satisfaction despite aleatoric uncertainty and bounded modeling errors.

Practical approaches include reachability analysis, which conservatively overapproximates all possible evolutions to guarantee safety~\citep{holmes2020reachable}, and control barrier functions (CBFs), which enforce invariance through real-time inequality constraints. When combined with control Lyapunov functions, CBF-based methods unify safety and goal achievement in an optimization framework, as demonstrated in tasks such as ball balancing~\citep{wang2025caging}. 

% When the bounds of the safe set are at risk of being violated, corrective actions are applied to keep the system within this set, thereby ensuring constraint satisfaction despite uncertainty. 
% Despite their promise, invariance-based methods remain underexplored in robot manipulation. The main challenge lies in contact, which introduces discontinuities in the dynamics, arises spontaneously and unpredictably, and is notoriously difficult to model. It complicates the design of invariance constraints and corrective controllers. A promising direction is to employ data-driven approaches to approximate invariant sets or to learn barrier functions directly from demonstrations and interaction data.

\subsubsection{Predictivity}
\label{control:predictivity}
Neuroscience suggests that human manipulation relies on the integration of long-term prediction with short-horizon reactivity~\citep{flanagan2006control}. Reactivity provides rapid sensory feedback, particularly tactile and visual, to detect mismatches between expected and actual outcomes and correct errors. Predictivity, on the other hand, enables the anticipation of collisions or contact events. In this way, predictive control primarily \emph{tolerates} uncertainty and variation by proactively compensating for their future effects, thereby helping \emph{prevent} failures before they occur. Purely local reactive control may fail in contact-rich settings, where abrupt changes in dynamics due to contacts can drive the system into unsafe states. To address this, model predictive control (MPC) incorporates a lookahead structure with reactive schemes. It optimizes actions over a finite horizon and executes the first control input. For example, MPC-based hybrid force–motion control is employed to simultaneously regulate end-effector trajectories and contact forces during interaction, explicitly accounting for contact mode transitions within the prediction horizon~\citep{jiang2025robust}; a contact-implicit MPC is utilized that predicts motion and contact forces within a trust region, enabling stable local control under unilateral contact and friction constraints~\citep{suh2025dexterous}; tube-based MPC guarantees that the real trajectory remains within a bounded ``tube'' around the nominal one, ensuring performance under external disturbances and model mismatch~\citep{nubert2020safe}. 

% Another direction is predictive neural controllers. Recent visuomotor policy learning also integrates receding-horizon control, improving action smoothness and robustness to latency~\citep{chi2023diffusion}. Furthermore, action chunking encodes short-horizon sequences of multi-step actions, providing foresight when used with temporal ensembles~\citep{zhao2023learning}.
% Zhao et al.~\citep{zhao2023learning} introduced action chunking, which predicts short-horizon chunks of actions and combines overlapping predictions through temporal ensembling. This design mitigates compounding errors and non-Markovian effects in human demonstrations.

% \begin{figure*}
%     \centering
%     \includegraphics[width=0.99\linewidth]{fig/learning-tax.pdf}
%     \caption{Taxonomy of mechanisms for learning robust manipulation policies.}
%     \label{fig:learning-tax}
% \end{figure*}

\begin{table*}[ht]
\centering
\caption{Overview of learning-based robustness mechanisms for robotic manipulation (Complementary to Table~\ref{tab:robustness_mechanisms}).
% \yifei{people mentioned "-centric robustness" may be redundant. any thoughts for alternative naming?}
}
\label{tab:learning_robustness_mechanisms}

\begin{tabular}{p{3.cm}p{6.1cm}p{5.1cm}p{1.2cm}}
\toprule
\textbf{Mechanism} & \textbf{Method} & \textbf{Principle} & \textbf{Section}\\
\midrule

\multirow{3}{*}{Training Distribution}
& Domain Randomization 
&  
& \multirow{3}{*}{\ref{learning-distribution-centric}}
\\
& Data Augmentation 
& Variation Toleration 
\\
& Data Scaling and Task Diversity 
& %; Temporary failure regulation 
\\
\midrule

\multirow{4}{*}{Policy Architecture}
& Perceptual and Structural Inductive Biases 
& Uncertainty \& Variation Toleration 
& \multirow{4}{*}{\ref{architecture-distribution-centric}}
\\
& Generative Policy Parameterizations 
& Uncertainty \& Variation Toleration 
\\
& Temporal Abstraction and Hierarchy 
& Failure Prevention %; Uncertainty\&variation toleration 
\\
& World-model-based Policies 
& {Uncertainty Reduction} % this is special 
\\
\midrule

\multirow{3}{*}{Learning Objective}
& Adversarial Training Objectives 
& Uncertainty \& Variation Toleration 
& \multirow{3}{*}{\ref{objective-distribution-centric}}
\\
& Safety- and Constraint-based Objectives 
& Failure Prevention 
\\
& Regularized and Conservative Objectives 
& Failure Prevention %; Uncertainty\&variation toleration 
\\
\midrule

\multirow{2}{*}{Policy Adaptation}
& Interactive Human-in-the-loop Imitation 
& \multirow{2}{*}{Temporary Failure Recovery }
& \multirow{2}{*}{\ref{adaptation-distribution-centric}}
\\
& Autonomous Continual Improvement 
& 
\\
% & Online Real-world RL 
% & Temporary failure regulation 
% \\
% & In-context and Prompt-based Adaptation 
% & Temporary failure regulation 
% \\

\bottomrule
\end{tabular}
\end{table*}
%%%%%%%%%%%%%%%%%%%%%%%%%%%%%%%%%%

\subsection{Policy Learning}
\label{policy_learning}
% Shaping/augmenting the training distribution
% Constructing the policy class
% Policy optimization objective
% Changing/adapting policy with episode distribution

% Coverage
% Inductive Bias
% Optimization Bias
% Adaptation

In parallel with the classical perception--planning--control pipeline, end-to-end policy learning has emerged as an alternative paradigm for manipulation. Learned policies can adapt to unstructured variations in the open world that are difficult to address analytically. In this sense, policy learning mainly improves robustness by \emph{tolerating} episodic variation and disturbances. Depending on the policy design, it can both \emph{prevent failure} through robust action selection and \emph{recover} after temporary task failure.

Within the formalism of Section~\ref{sec:formulation}, these methods instantiate a parameterized policy~$\pi_\phi$ and train it from data so that the robust performance functionals~$\Gamma(\pi)$ remain high across an episodic distribution $\rho_\mathcal{N}(\nu)$. In the rest of this section, we group policy-learning approaches according to four complementary ways (Table~\ref{tab:learning_robustness_mechanisms}) in which they achieve robustness: some methods shape the tasks, environments, and perturbations that $\pi_\phi$ is trained on (\ref{learning-distribution-centric}); some methods design policy classes and architecture whose inductive biases and temporal structure reduce sensitivity to nuisance variation and partial observability (\ref{architecture-distribution-centric}); some methods modify the learning objective to better align training with robust performance (\ref{objective-distribution-centric}); and others allow policies to adapt at deployment through residual corrections, online updates, or test-time adaptation (\ref{adaptation-distribution-centric}).

\subsubsection{Training Distribution}
\label{learning-distribution-centric}
In the notation of Section~\ref{sec:formulation}, robustness is measured by the functional~$\Gamma(\pi)$ evaluated over an episodic distribution $\rho_\mathcal{N}(\nu)$. A branch of approaches acts directly on this distribution: it shapes how training episodes are generated while keeping the policy class $\Pi$ and the learning objective fixed. These methods aim to \emph{tolerate} variations across episodes by broadening the training distribution over dynamics, initial states, environments, and goals.

A first family of methods uses \emph{domain randomization}, widely used, particularly in reinforcement learning. Instead of a single nominal environment, one defines a family of ``worlds'' $\theta$ (and sometimes $x_0$ and $\mathcal{G}$) from a broad distribution. Policies trained under such variation can transfer to real manipulation with minimal tuning, as demonstrated for vision-based grasping appearance~\citep{tobin2017domain} and for dexterous in-hand manipulation~\citep{akkaya2019solving,andrychowicz2020learning}. 
By expanding the training distribution $\rho_\mathcal{N}$ to encompass diverse scenarios, these methods enable the policy to \emph{tolerate} variations in dynamics and appearance.

A second line achieves robustness via \emph{data augmentation} on training trajectories. These methods keep the generator of $\nu$ fixed, but expand the empirical distribution $\hat{\rho}_\mathcal{N}$ by transforming demonstrations according to task priors: prior work applies SE(2)/SE(3) transformations to scenes and actions to exploit invariance and equivariance~\citep{florence2019self, mandlekar2023mimicgen, ameperosa2025rocoda}, re-renders fixed trajectories from novel viewpoints to vary appearance~\citep{zhou2023nerf, zhang2024diffusion}, and retargets demonstrated trajectories to new object and scene configurations~\citep{yu2023scaling, chen2023genaug, xue2025demogen}. Beyond such structured priors, a related class injects calibrated noise into states and actions to mitigate covariate shift and compounding error, making the policy more robust~\citep{laskey2017dart, ke2021grasping, simchowitz2025pitfalls}. 

A third class relies on \emph{data scaling and task diversity}, as most visibly demonstrated in recent vision--language--action (VLA) models. Rather than explicitly parameterizing randomization or augmentations, these methods assemble very large, heterogeneous datasets of manipulation episodes spanning many robots, scenes, tasks, and language-specified goals, and train a single policy across this mixture~\citep{brohan2022rt, zitkovich2023rt, o2024open, kim2024openvla}. Robustness here is treated as generalization over a very broad empirical $\hat{\rho}_\mathcal{N}$ that pools demonstrations and deployments across embodiments, tasks, and environments, with a single $\pi_\phi$ expected to maintain high $\Gamma(\pi_\phi)$ throughout this mixture.
An important direction in data scaling extends beyond successful task executions to include \emph{recovery} behaviors following \emph{temporary failures}, capturing the adaptive strategies required in unstructured environments. Policies trained only on ideal trajectories may overfit to nominal conditions. For example, Large Behavior Models incorporate hybrid sim-to-real datasets containing both successful executions and recovery episodes~\citep{barreiros2025careful}.

\subsubsection{Policy Architecture}
\label{architecture-distribution-centric}
A policy is a mapping $\pi : \mathcal{B} \to \mathcal{U}$ from beliefs to actions. A line of approaches pursues robustness by designing policy architecture and data representation: they constrain how information about observations is encoded and how actions are parameterized, without changing the training distribution $\rho_\mathcal{N}$ or the learning objective. The goal is to make the mapping $b_t \mapsto u_t$ inherently \emph{tolerant} to nuisance variation, partial observability, and long horizons.

First, \emph{perceptual and structural inductive biases} build a geometric state from raw observations. Keypoint-based representations encode objects and goals as a small set of 2D/3D landmarks and express tasks as geometric relations among these keypoints, so the policy operates in a low-dimensional space aligned with contacts and affordances rather than raw pixels~\citep{manuelli2019kpam, qin2020keto, huang2024rekep}. SE(3)-equivariant representations constrain internal features (and sometimes actions) to transform consistently under rigid motions of the workspace, so that rotating or translating the scene induces the same transformation in the encoded state and allows one controller to reuse a strategy across pose-shifted instances of a task~\citep{simeonov2022neural, ryu2022equivariant, eisner2024deep, wang2024equivariant}. Graph-structured encoders treat objects, robot links, and goals as nodes with edges capturing spatial or relational constraints, and use message passing to learn local interaction rules that generalize to scenes with more objects, new goal configurations, and multi-object rearrangement~\citep{li2020towards, lin2022efficient, huang2022planning}. Across these designs, robustness comes from aligning the learned representation with the geometry and relational structure of the task, so that pose- and relation-preserving changes in the scene tend to yield similar decisions, resulting in stable performance under these variations. 

Second, \emph{generative policy parameterizations} implement $\pi_\phi$ as a conditional generative process over short action sequences, most commonly using diffusion or flow-matching models for manipulation tasks~\citep{chi2023diffusion, black2410pi0}. The policy maps beliefs $b_t$ to actions by starting from injected noise and iteratively refining a sample via a supervised denoising or flow-integration process. Empirically, this iterative computation with noise injection induces an inductive bias that improves closed-loop robustness to covariate shift~\citep{pan2025much}. 

\emph{Temporal abstraction and hierarchy} represent actions as skills or options that span multiple timesteps. The policy acts in an extended action space of options: a high-level policy selects a skill, and a low-level controller executes it until termination, so the decision-making problem becomes a semi-MDP with a shorter effective horizon~\citep{sutton1998intra}. Hierarchical RL methods learn both the skills and the high-level policy, often using subgoal states as the interface between levels and off-policy updates to remain sample efficient in continuous control~\citep{nachum2018data}. In manipulation, skill-based approaches learn and plan with such temporally extended behaviors, such as predicting long-horizon discrete action sequences from a scene image for sequential physical reasoning~\citep{driess2021learning}, discovering reusable closed-loop skills from unsegmented demonstrations and training a meta-controller to compose them~\citep{zhu2022bottom}, or chaining separate diffusion models for individual skills to solve unseen long-horizon tasks under constraints~\citep{mishra2023generative}. Robustness in this class arises from horizon reduction: the high-level policy makes skill-level decisions per episode, which reduces compounding error, supports reuse of skills across task variations, and thereby helps \emph{prevent} failures before they occur.

Lastly, \emph{world-model-based policies} couple the controller with a learned dynamics model in a latent space. They maintain a latent state $z_t = e(o_{0:t}; \theta)$ together with a predictive model $\hat{p}_\psi(z_{t+1} \mid z_t, u_t)$, and use short imagined rollouts in this latent space to reason about future states before committing to an action~\citep{hafner2019learning, ichter2019robot, hansen2022temporal, hafner2023mastering}. World-model-based policies can \emph{reduce} effective epistemic uncertainty by learning predictive structure over how the environment evolves under the robot's actions. Rather than eliminating uncertainty, the learned model provides an internal approximation of otherwise unknown or partially observed dynamics that can be queried during decision making. In manipulation, such latent models are used for planning and runtime monitoring to score candidate actions under many predicted futures and reject those that would enter unsafe, constraint-violating, or high-risk regions~\citep{liu2024model, nakamura2025generalizing, sun2025latent}.

\subsubsection{Learning Objective}
\label{objective-distribution-centric}
While distribution- and architecture-centric approaches shape what the policy sees and how it represents it, in this part we introduce methods that instead shape what the policy optimizes for, embedding worst-case performance, safety, or conservatism directly into the learning objective.

\emph{Adversarial training objectives} make robustness explicit by optimizing the min--max criteria of Eq.~\eqref{eq:robust-policy-maxmin} rather than a nominal expected return. The policy is trained against an adversary that chooses disturbances, environment parameters, or competing behaviors to minimize task performance while the protagonist maximizes it. In practice, the adversary injects destabilizing forces or physical perturbations to grasps and object poses, so policies are trained on deliberate worst cases rather than randomly sampled disturbances~\citep{pinto2017robust,pinto2017supervision,jian2021adversarial}. By repeatedly optimizing against these hard cases, the learned policy becomes less sensitive to uncertainty and variation, thereby improving its ability to \emph{tolerate} them at deployment.

\emph{Safety- and constraint-based objectives} restrict which policies are admissible when we evaluate robustness. In the POMDP view of Section~\ref{sec:formulation}, one augments the control problem with constraint costs and searches for policies that achieve high task performance $\Gamma(\pi)$ while keeping these costs below a threshold, as in constrained MDP methods such as Constrained Policy Optimization~\citep{achiam2017constrained}. Related work uses safety layers or control barrier functions that project or override the learned action when it would violate certified constraints~\citep{dalal2018safe,cheng2019end}, thereby \emph{preventing} failures by keeping execution within admissible regions. 

\emph{Regularized and conservative objectives} aim to improve robustness to epistemic uncertainty and limited data coverage by constraining how far the learned policy and value function extrapolate beyond the empirical episodic distribution~$\hat{\rho}_\mathcal{N}(\nu)$. Offline RL work frames this as ``extrapolation error'' and shows that robust performance typically requires (i) behavior regularization, which keeps $\pi$ close to actions well supported by the data, and (ii) pessimistic value estimates that down-weight out-of-distribution actions~\citep{levine2020offline,kumar2020conservative,kostrikov2021offline,fujimoto2021minimalist}. In real-robot manipulation, these ideas appear in conservative offline RL that improves over imitation on tabletop tasks using logs from safe operation~\citep{zhou2022real}, in large-scale deployments that rely on heavily regularized off-policy updates for stable long-term waste-sorting with mobile manipulators~\citep{herzog2023deep}, and in regularized imitation methods that combine optimal-transport trajectory matching with a pull toward demonstration behavior to achieve few-shot real visual manipulation~\citep{haldar2023watch}. Across these examples, robustness comes from biasing learning toward regions of state–action space that are well covered and conservatively valued, thereby reducing out-of-distribution actions and helping \emph{prevent} failures during execution.

\subsubsection{Policy Adaptation}
\label{adaptation-distribution-centric}
This line of work assumes a base policy $\pi_{\text{base}}$ has been trained under some episodic distribution $\rho_\mathcal{N}^{\text{train}}(\nu)$, and focuses on how its behavior is modified using feedback from the deployed environment when the test-time distribution $\rho^{\text{test}}(\nu)$ differs from $\rho_\mathcal{N}^{\text{train}}(\nu)$. Unlike distribution-centric approaches, they do not redesign $\rho_\mathcal{N}$ up front, but instead close the loop between deployment performance and further adaptation. 

First, \emph{interactive human-in-the-loop imitation} refines a pretrained policy at deployment using online expert feedback. Humans can intuitively identify and correct failure-imminent situations or recover from temporary \emph{failure}, providing targeted supervision where pre-trained policies are weakest~\citep{liu2022robot}. In DAgger-style methods, imitation learning is reduced to online no-regret learning: at each iteration, the current policy is rolled out to generate states, the expert labels those visited states with preferred actions, these pairs are added to an aggregate dataset, and a new policy is trained on the union~\citep{ross2011reduction}. This ensures that the learned stationary policy performs well under the state distribution it induces, rather than only on states seen in expert demonstrations, which directly addresses covariate shift in sequential decision making. Subsequent work adapts this idea to human experts in safety-critical or latency-limited settings~\citep{kelly2019hg}, to remote teleoperation for contact-rich manipulation~\citep{mandlekar2020human}, to ``on-the-job'' deployment where autonomy and learning run continuously with humans stepping in on hard cases~\citep{liu2022robot}, and to kinesthetic delta corrections with force-aware residual policy via compliance~\citep{xu2026compliant}.

Second, \emph{autonomous continual improvement} adapts policies during deployment without relying on explicit expert labels. One branch uses online RL on the real robot: a warm-start policy (often from demonstrations or offline data) continues to collect rollouts on the hardware, and policy updates $\pi_\phi$ under the actual test-time episodic distribution, with mechanisms such as safety constraints, automatic or cheap resets, and simple reward specification to make this practical for contact-rich manipulation~\citep{levine2016end, rajeswaran2017learning, johannink2019residual, xu2022dexterous, luo2024serl, luo2025precise}. A second branch uses {in-context and prompt-based adaptation}: sequence models are trained on multi-task data so that, at test time, a short prompt of teleoperated demonstrations or recent trajectories for the current variation $\nu$ conditions the model, enabling few-shot adaptation to new objects, layouts, and goals without gradient updates~\citep{duan2017one,valassakis2022demonstrate,fu2024context}.

\subsection{Hardware}
Manipulation robustness can also be achieved through the robot's embodiment, complementing the contributions of perception, planning, control, and learning. This perspective is grounded in the concept of \emph{morphological computation}, where physical structures perform functions that would otherwise require explicit control~\citep{paul2006morphological}. Through the design of materials, geometry, and actuation, the body can shape the system dynamics~$f$ such that a wider range of states~$x_t$ and actions~$u_t$ lead to successful outcomes, while reducing the consequences of errors in perception, modeling, and execution.

A recurring theme of hardware-based robustness is the \emph{toleration} of object variation and external disturbances through physical design. This can be achieved through passive compliance (Section~\ref{mechanical:passive-compliance}), which absorbs disturbances and accommodates contact uncertainty, or through adhesive contact mechanisms (\ref{mechanical:adhesion}), which remain effective despite geometric variation and perturbations. In addition, some robotic systems achieve robustness through \emph{morphological adaptation} (\ref{mechanical:morphology-adaptation}), reconfiguring their physical structure in response to changing conditions and thereby restoring performance when an existing morphology becomes ineffective.

% More specifically, mechanical intelligence often improves robustness by designing the environment/robot parameter~$\theta$ (e.g., joint compliance, stiffness, contact properties), which directly changes~$f$. The same controller or policy thereby achieves better performance~$\Gamma(\pi)$ under uncertainty and variation.
% or operation modes across episodes/tasks to keep objectives tractable under changing conditions 
% passive compliance: shaping local impedance at contacts, 
% adhesion: modifies contact properties and the feasible wrench set 
% \emph{postural synergy} constrains the reachable configuration or action space to low-dimensional subspaces (\ref{mechanical:synergy}); 
% sota mechanical intelligence weakly connected to other subfields - explore more like co-design
% For example, octopuses exploit the compliance of their arms to naturally conform around objects, achieving stable grasps without precise motion planning. 
% By offloading computation from the brain to the body, robustness can emerge naturally through simpler control strategies. Such mechanical intelligence provides inherent tolerance to uncertainties and variations in the environment and thus prevents failure. 

\subsubsection{Passive Compliance}
\label{mechanical:passive-compliance}
Passive compliance enhances manipulation robustness by physically \emph{tolerating} disturbances and contact uncertainty through compliant, soft surface material, before they propagate through the control loop. It can be viewed as a mechanical counterpart to active compliance (Section~\ref{control:compliance}) that shapes the interaction dynamics prior to control. By allowing bounded motion in response to external forces, compliant structures reduce the sensitivity of contact interactions to disturbances, geometric variation, and modeling errors. As a result, reliable and adaptive manipulation can often be achieved with reduced reliance on precise sensing, accurate models, and high-bandwidth control.

\begin{figure}
    \centering
    \includegraphics[width=0.99\linewidth]{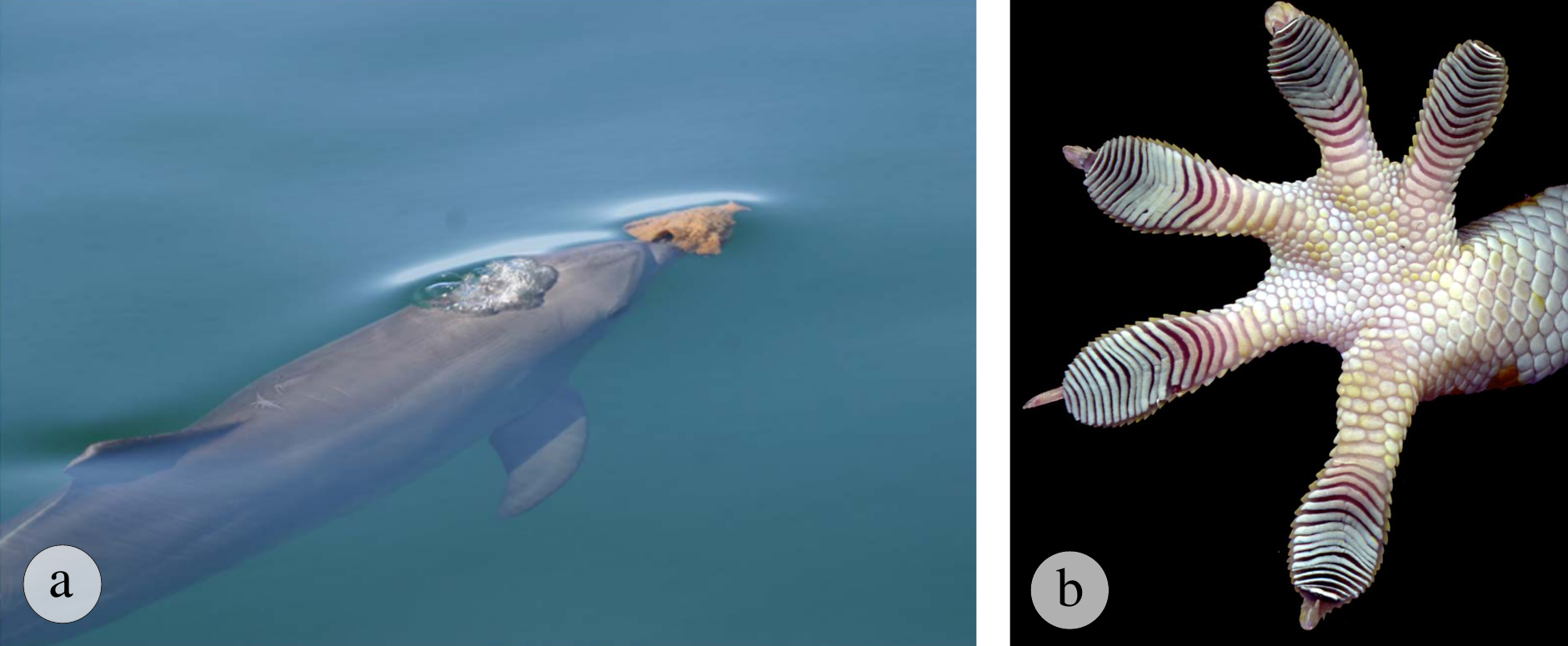}
    \caption{Examples of hardware mechanisms for robust manipulation.
    Hardware intelligence in robotics is often inspired by biological strategies. 
    (a) Passive compliance: Dolphins wrap marine sponges around their beaks for protection~\citep{mann2008dolphins}, while robotic hands benefit from compliant surface material.
    % in finger joints, wrists and skin~\citep{junge2025spatially} (\copyright\ Springer Nature, adapted under CC BY).
    (b) Adhesion: gecko feet inspire adhesive grippers, such as~\cite{song2017controllable}.
    Figure (a) is adapted under CC BY; image courtesy of (b): Prof. Kellar Autumn, Lewis and Clark College~\citep{autumn2006tokayfoot}.
    % (c) Postural synergy: elephants control their trucks with muscular synergies~\citep{dagenais2021elephants} (\copyright\ Elsevier), a principle adopted in adaptive robotic hands~\citep{catalano2014adaptive} (\copyright\ SAGE).
    % (c) Morphology adaptation: a robotic gripper with reconfigurable fingers, adjusted by rigid rods (ii), adapts its geometry to target objects~\citep{pagoli2021soft} (\copyright\ IEEE).
    % \yifei{remove c.}
    }
    \label{fig:hardware}
\end{figure}

This principle is widely observed in both biological and robotic systems. Dolphins, for example, use marine sponges as protective tools while foraging on rocky seabeds (Fig.~\ref{fig:hardware}-a), thereby tolerating environmental uncertainty and avoiding damaging contacts~\citep{mann2008dolphins}. In robotic manipulation, passive compliance is commonly used to {tolerate} misalignment and reduce impact forces in assembly tasks~\citep{drake1978using}. Similar principles underlie soft grippers based on compliant materials or tendon-driven structures, which passively conform to object geometry under uncertainty~\citep{deimel2013compliant}. Hybrid designs further combine compliance with controllable stiffening, as exemplified by granular-jamming grippers, enabling compliant exploration followed by adaptive stiffening to securely grasp the object~\citep{shintake2018soft,amend2012positive}.

% The Remote Center Compliance device introduces compliant motion in selected directions to absorb positional offsets and prevent jamming during peg-in-hole assembly~\citep{drake1978using}
% Similarly, Series Elastic Actuators place a spring between the motor and the load, providing shock absorption and energy storage~\citep{pratt1995series}.
% controlled stiffness, which refers to the use of materials or mechanisms that can switch between a soft, compliant state and a stiff state during manipulation. For example,
% However, flat or highly deformable objects, such as fish fillets, cannot be reliably grasped with this strategy, as excessive pressure during stiffening may damage delicate items~\citep{shintake2018soft}.
% This allows stable grasping without the need for precise sensing or control.
% such as granular jamming, low-melting-point alloys, or shape-memory polymers

\subsubsection{Adhesion}
\label{mechanical:adhesion}
Adhesive material modifies robot-object contact surfaces to sustain strong tangential (shear) forces with minimal normal force, which is particularly suitable for delicate, thin, or flat objects. 
By enabling surface-based attachment rather than relying solely on friction-limited point contacts, adhesion alters the contact dynamics and makes manipulation more \emph{tolerant} to uncertainty in contact location, object pose, and contact modeling. Common adhesion mechanisms include gecko-inspired dry adhesion and vacuum-based suction~\citep{shintake2018soft}.
Gecko-inspired adhesives exploit van der Waals forces to generate strong shear interactions across a wide range of surface textures without requiring active control (Fig.~\ref{fig:hardware}-b). Such systems have been used to grasp bulky objects in microgravity~\citep{jiang2017robotic}, demonstrating robustness to object misalignment and uncertain contact conditions. Vacuum-based suction is another widely adopted adhesion mechanism. Besides \emph{adhesion}, many suction cups incorporate bellows that provide \emph{passive compliance}, allowing the system to accommodate pose errors and conform to local surface curvature. Multi-affordance manipulators that combine suction cups with parallel-jaw grippers~\citep{zeng2022robotic} further exploit mechanical \emph{redundancy}, improving robustness and extending applicability across a broader range of objects and environments.

\subsubsection{Morphology Adaptation}
\label{mechanical:morphology-adaptation}
The environment and task requirements encountered by a robot are often not fixed. As a result, a morphology that performs well under one set of conditions may become ineffective when the environment, object properties, or task requirements change. Some robotic systems address this challenge by adapting their embodiment, modifying geometry, stiffness distribution, or kinematic structure to suit new conditions. In this way, morphological adaptation can be viewed as a mechanism for recovering from \emph{temporary failure} caused by changing environments or task requirements~\citep{liang2026modular}. Complementary to passive compliance and adhesion, which shape interactions locally at the contact interface, morphological adaptation reshapes the embodiment at a global level.

In soft robotic hands, for example, the bending location and range can be adjusted by inserting a stiff rod into the center of each finger~\citep{pagoli2021soft}. Such reconfiguration enables the end-effector to switch between grasping modes and maintain performance across different object types without requiring complex sensing or control. PolyBot~\citep{yim2000polybot} further illustrates \emph{modular} reconfiguration, where detachable modules autonomously rearrange to accommodate new workspace constraints. By adapting their morphology to changing conditions, such systems improve robustness, support fault tolerance, and extend manipulation capabilities in unstructured environments.

% , beyond adapting the robot parameterization $\theta$ given an embodiment. This effectively selects a morphology that makes the same high-level objective more tractable under changing episodes. 
% At smaller scales, reconfigurable microrobot swarms~\citep{xie2019reconfigurable} can transition between collective modes (e.g., liquid, chain, vortex) by tuning external magnetic fields, allowing the system to adapt its interaction mode to fluidic disturbances while preserving task intent.
% Together, these examples illustrate how reconfigurability across scales enhances manipulation robustness by allowing robots to physically adapt to environmental variability and task demands.
% traverse confined spaces or
\section{Evaluation Methods of Manipulation Robustness}
\label{sec:evaluation}

Evaluating manipulation robustness requires quantitative criteria that reflect a system’s capability to achieve its task objectives under challenges. In this section, we first present \emph{empirical} evaluation protocols that assess how effectively a robot performs physical manipulation tasks in simulation or in the real world. We then revisit \emph{analytical} evaluation protocols that assess robustness using grasp-quality measures, margin-based metrics, and other abstract mathematical constructs rather than empirical test-time task performance.

\subsection{Empirical Evaluation Protocols}
Based on task performance, these evaluation protocols empirically assess whether and how effectively a manipulator achieves the task goal~$\mathcal{G}$ despite varying challenges. They provide more direct comparisons for end users~\citep{liconti2026benchmark} and are widely adopted by both empirical (learning-based) and analytical methods.

\subsubsection{Goal-Oriented Measures}
Given a manipulation task specification together with domains of epistemic uncertainty, aleatoric uncertainty, and episodic variation, the most widely used robustness evaluation protocol is to measure empirical task success or failure~\citep{mason2018toward}. This approach, commonly referred to as \emph{Monte Carlo} robustness tests, executes the system repeatedly under a sampled model mismatch $(\hat{\theta},\theta)$, stochastic disturbances $w_t$ and $v_t$, and episodic variations $\nu$, and estimates the probability of achieving the task goal. Concretely, robustness is the expected probability of goal satisfaction under the challenges,
\begin{equation}
    % \mathbb{E}_{\nu \in \mathcal{N},\, w_t \in \mathcal{W},\, v_t \in \mathcal{V}}
    \mathbb{E}_{\nu \sim \rho_\mathcal{N}, w_{0:T} \sim \rho_\mathcal{W}, v_{0:T} \sim \rho_\mathcal{V}}
    \left[\mathbf{1}\{x_T \in \mathcal{G}\} \mid (\hat{\theta}, \theta) \right],
\label{eq:evaluation-target}
\end{equation}
often approximated by empirical success rates ${n_{\text{success}}}/{n_{\text{total}}}$ computed over multiple trials. This evaluation directly aligns with the robustness formulations introduced in Section~\ref{sec:formulation}.

Following this evaluation protocol, examples include success rates in object pose estimation under perception noise and occlusion~\citep{mandlekar2023mimicgen}; success rates or expected returns in reinforcement learning, which often reduce to goal satisfaction indicators, evaluated across episodic variations such as different initial object placements or object instances~\citep{luo2025precise,andrychowicz2020learning}; and success under adversarial perturbations, including external disturbances applied to manipulated objects or injected joint-level perturbations during execution~\citep{jian2021adversarial}. Variants of this approach include {systematic parameter sweeps}, where parameters in $\theta$ (e.g., inference latency, friction coefficients, lighting conditions) are varied across controlled ranges and performance is plotted as a function of the parameter~\citep{chi2023diffusion}.

\subsubsection{Stage-wise Measures}
While the formulation in Eq.~\eqref{eq:evaluation-target} focuses on terminal goal satisfaction, there are many manipulation tasks more naturally characterized by \emph{trajectory-level subgoals}. In such cases, success depends not only on reaching a final state but also on correctly executing a sequence of intermediate stages. Consequently, several works adopt stage-wise evaluation metrics, in which a task is decomposed into semantically meaningful phases and success is assessed at each stage. For example, \cite{kang2026learning} evaluate fragile-object manipulation policies across three stages: approach, stable grasp, and task completion. \cite{xue2025reactive} decompose a paper-cup lifting task into clamping and lifting stages. 
Evaluators may also design task-specific semantic rubrics to assess policy behavior beyond aggregate success rates~\citep{kress2024robot}. For instance, in a \emph{Flip-and-Serve Pancake} task, relevant criteria may include: ``Robot collided with anything?'', ``Robot flipped pancake?'', ``Robot picked up pancake?'', in addition to overall task success.
The stage-wise protocols provide a more fine-grained assessment of policy performance, enabling the identification of failure modes that may be obscured by a single binary success measure.

% \begin{figure}
%     \centering
%     \includegraphics[width=0.9\linewidth]{fig/survey-evaluation.pdf}
%     \caption{Examples of robustness evaluation methods. Margin to failure: coffee inside a mug has a safety margin below the rim of the mug (Adapted from \citep{sternad2016predictability}, under the Creative Commons Attribution license).
%     % (b) Convergence metrics: self-correcting behaviors can be exploited in the system dynamics using convergence metrics~\citep{kong2019optimally} (\copyright\ Springer).
%     }
%     \label{fig:evaluation}
% \end{figure}

\subsection{Analytical Evaluation Protocols}

While success probability under challenges provides an empirical, task-level notion of robustness, it is often insufficiently fine-grained for answering more specific questions about \emph{how} a manipulation system withstands challenges. In many scenarios, one is not only interested in whether a task succeeds, but also in the structure and degree of tolerance to disturbances, uncertainty, or variation. This is where analytical evaluation protocols become valuable: they aim for potential capability, rather than only realized capability, through explicit mathematical constructs.
They are structured and interpretable, tailored to specific manipulation mechanisms and failure modes. 
They are often approximated by the success probability in Eq.~\eqref{eq:evaluation-target}.
Unlike performance-based protocols, which primarily serve as empirical \emph{diagnostic tools} for robustness assessment, analytical protocols can also function as \emph{optimization objectives} embedded within planning, control, or learning methods for the synthesis of robust manipulation behavior.
As follows, we introduce several analytical evaluation protocols. 

\subsubsection{Closure-based Grasp Quality Measures}
Closure properties characterize how contact forces and geometric constraints prevent an object from deviating from a desired configuration or set of configurations. These properties are typically analyzed from two different perspectives: \textit{local} conditions at the contact and \textit{global} conditions within the object's configuration space.

Local measures derived from force closure or form closure quantify the instantaneous robustness of quasistatic manipulation via wrench-space or geometric analysis. These measures can be evaluated either analytically or empirically. Analytical approaches, commonly referred to as grasp quality measures~\citep{ferrari1992planning}, compute robustness using physics-based models to determine the system's ability to resist external disturbances. In contrast, empirical approaches approximate these analytic measures by learning from data. For instance, grasp datasets labeled with analytic quality values are used to train neural networks that predict robustness directly from visual or sensory inputs~\citep{mahler2017dex}, enabling scalable estimation in unstructured environments.

While local measures assess stability via instantaneous force or geometric conditions, many complex manipulation tasks depend on the global characteristics of the configuration and contact spaces—specifically, how reachable, connected, or ``trapped'' an object remains within its environment. Topological analysis captures such global invariants, providing a perspective complementary to purely local evaluations~\citep{pokorny2015data, bhattacharya2015persistent}. For example, energy-bounded caging~\citep{mahler2018synthesis} defines robustness in terms of the size or persistence of the configuration-space region that keeps the object contained. In this context, a larger connected component implies a greater structural tolerance to uncertainty or disturbances.

\subsubsection{Margin-based Measures}
\label{sec:eval-margin}
``Margin'' to failure quantifies robustness by measuring the distance between a system’s operating state and the boundary of task failure. Intuitively, systems maintaining larger margins are more robust, as they can tolerate greater disturbances, modeling errors, or stochastic noise before a failure occurs. Consider the task of placing a cup of coffee on a coaster: precise positioning is secondary as long as the liquid level remains safely below the rim. If the goal region~$\mathcal{G}$ is defined as the set of states where all liquid remains inside the cup, its complement $\mathcal{G}^{c}$ corresponds to failure (spillage). While a binary reward only detects the occurrence of failure, a safety margin quantifies the resilience of the current state:
\begin{equation}
\sigma_{\text{margin}} = \operatorname{dist}(x, \partial \mathcal{G}),
\end{equation}
where $\operatorname{dist}(\cdot)$ denotes a task-appropriate distance metric and $\partial \mathcal{G}$ is the boundary of the goal region. Positive values indicate the system is within $\mathcal{G}$, with larger values implying greater robustness.

In neurophysiology, this distance is often defined through energy, i.e., the energy required for the system to transition from a safe state $x$ to failure $\mathcal{G}^c$~\citep{hasson2012energy}. Humans intuitively maintain such margins by relying on simplified internal models rather than exact dynamics, selecting control strategies that preserve safety margins~\citep{sternad2016predictability}. Similar concepts have been applied to evaluate robotic manipulation robustness~\citep{dong2024characterizing}. Analogous ideas appear in force- and wrench-based robustness metrics, such as grasp stability measures that quantify the maximum external wrench a grasp can resist before slippage~\citep{roa2015grasp}. Extending the concept over time leads to the notion of a ``safety tube'', a region surrounding a nominal trajectory within which the system can remain despite bounded disturbances~\citep{fox2006exploration,nubert2020safe}.

\subsubsection{Signal Temporal Logic Measures}
Signal Temporal Logic (STL) is a formal language for describing time-dependent task requirements over real-valued signals, such as contact force, object position, or velocity — examples include conditions such as maintaining a grasp force below a threshold, reaching a target region within a given time, or keeping an object stable throughout execution. STL further provides quantitative semantics that assign a robustness score measuring how strongly a trajectory satisfies or violates the specified requirements~\citep{maler2004monitoring, donze2010robust}. In robotics manipulation, STL robustness has been predominantly used as an optimization target — either as a reward signal in reinforcement learning~\citep{kapoor2020model, li2017reinforcement}, a loss function for neural predictive control~\citep{meng2023signal}, or a planning objective in task-and-motion planning~\citep{takano2021continuous}. Its use as a post-hoc evaluation metric remains rare: \cite{kress2024robot} have explicitly leveraged STL robustness for grading learned manipulation policies after training, demonstrating how the robustness score reveals not only whether a policy succeeds but how close it is to failure across complex, temporally structured objectives. The STL measures are conceptually related to the margin-to-failure measures discussed above, but generalize them to composite temporal specifications involving sequencing, timing, and conditional constraints. A practical limitation is that STL robustness is scale-dependent: predicates expressed in different physical units (e.g., millimeters versus Newtons) produce robustness values on incomparable scales, so that the min/max aggregation inherent in the semantics can be dominated by the choice of units rather than by task-relevant difficulty~\citep{varnai2020robustness, dhonthi2021study}. Despite these challenges, the broader use of STL measures as an evaluation tool for manipulation robustness remains a promising direction.

\subsubsection{Convergence and Divergence Measures}
In dynamically complex manipulation tasks such as tossing, catching, or balancing, robustness often arises from the intrinsic structure of the system dynamics. Humans routinely exploit such dynamics by inducing self-correcting behaviors in which small perturbations are naturally compensated, allowing motion to remain stable under bounded aleatoric disturbances $w \in \mathcal{W}$. From an evaluation perspective, this form of robustness can be captured by convergence or divergence measures~\citep{bazzi2020robustness}, which quantify how trajectories evolve under perturbations. For example, the largest eigenvalue of the symmetric part of the system Jacobian measures the local rate of trajectory divergence or contraction, with negative values indicating stabilizing, disturbance-rejecting dynamics. Such measures can be incorporated into constrained optimization or planning formulations, e.g., as constraints on contact dynamics, to favor actions that induce convergent behavior and naturally funnel executions back toward desired outcomes.

\section{Discussions}
\label{sec:discussion}
In this section, we discuss several complementary insights and implications that emerge from the robustness mechanisms.

% \section{Insights and Takeaways}
% We focus on cross-cutting principles underlying manipulation robustness, strategies for managing failure, and broader perspectives and takeaways for practitioners.
% \yifei{perception, planning - both reduction and toleration; hardware mainly toleration? any deeper reasons}

\subsection{Robustness and Performance Trade-offs}
Manipulation performance is inherently context-dependent. A system may exhibit high performance in controlled laboratory settings, yet experience sharp degradation in unstructured real-world environments. Systems optimized for narrow conditions can achieve exceptional speed or precision, but often do so at the expense of robustness when those conditions change. In nature, star-nosed moles exemplify this trade-off: their specialized nasal appendages enable rapid prey detection in wetland tunnels, but such specialization may be fragile under environmental shifts~\citep{catania1996unusual}. By contrast, some systems with moderate peak performance may trade efficiency or accuracy for robustness. Compliant robotic grippers, for example, adapt well to variations in object shape or mass, but typically sacrifice positional precision~\citep{deimel2013compliant}. These observations suggest that progress in manipulation should be evaluated not only by peak performance under ideal conditions, but by the ability to sustain reliable function under imperfect and unpredictable in-the-wild challenges.

\subsection{Robustness and Related Concepts}
Robustness often intersects with but differs from other fundamental concepts, including safety, stability, generalizability, etc. \textit{Safety} concerns preventing harm to humans, the environment, or the robot itself, and is critical in domains such as autonomous driving, navigation, and human–robot interaction~\citep{brunke2022safe}. \textit{Stability}, common in control and grasping, refers to maintaining or converging to a desired state under perturbations. Robustness extends beyond these notions, encompassing not just safe or stable operation but the broader pursuit of general task goals under uncertainty or variation. 
% \yifei{to be updated.}
% Robustness and \textit{generalizability} are frequently discussed interchangeably, even though they capture different aspects of system behavior.
% Generalization refers to a learned model’s ability to handle out-of-distribution data or novel domains, whereas robustness emphasizes measurable and reliable performance within the domain despite uncertainty and variation. 

Robustness and \textit{generalizability} are also frequently discussed interchangeably, especially in robot learning, but they capture different aspects of system behavior. Generalizability concerns the scope over which a learned policy or model can be applied beyond its training conditions, such as new objects, scenes, embodiments, task instructions, or simulation-to-real transfer~\citep{kroemer2021review,aljalbout2025reality,gao2026taxonomy}. Robustness, in contrast, concerns the reliability of task achievement under a specified set or distribution of uncertainty and variation. Thus, generalization may serve as one mechanism for robustness when the deployment variations are covered by the generalized competence of the policy, but it is neither necessary nor sufficient: a controller may be robust within a narrow operational condition without generalizing broadly, while a generalist policy may cover many tasks yet remain sensitive to small perturbations, contact uncertainty, or distributional shifts not represented in its evaluation protocol.

\subsection{Insights from beyond Manipulation Robustness}
Robustness research in domains beyond manipulation offers valuable insights. As mentioned in Section~\ref{sec:definition}, locomotion robustness is particularly relevant, as locomotion can be viewed as a form of ``self-manipulation''~\citep{johnson2016hybrid,mason2018toward}. Techniques such as adversarial training for controller robustness~\citep{shi2024rethinking}, widely explored in locomotion, remain comparatively underutilized in manipulation and represent a promising direction. Similarly, biological inspirations such as gecko-inspired adhesives, originally studied in climbing and locomotion, have motivated gripper designs~\citep{jiang2017robotic}. Despite these overlaps, manipulation presents unique challenges: unlike locomotion, which often reduces interaction to discrete foot--ground contacts, manipulation involves rich, multi-body interactions among the hand, object, and environment, with diverse contact types and combinatorial complexity that significantly complicate robustness analysis and design.

\subsection{Beneficial Role of Perturbations}
Finally, perturbations, typically viewed as detrimental, can in some cases enhance robustness. In granular media, small vibrations can break clogging arches when pouring sand into a funnel~\citep{to2001jamming}. The vibrating bowl feeder discussed in Section~\ref{subsec:planning-funnel} provides another example, where controlled vibrations guide parts toward consistent configurations. Analogously in manipulation, slight, intentional perturbations can dislodge unstable contacts, reduce excessive contact forces, or guide objects out of shallow local minima. Everyday actions such as wiggling a Jenga block or flicking an omelette exemplify how controlled perturbations stabilize interactions rather than destabilize them. These observations highlight that robustness is not always achieved by tackling uncertainties or variations, but sometimes by strategically exploiting them.

% % % % % % % % % % % % % % % % % % % % % % % % % % % % % % % % % 
% % % % % % % % % % % % % % % % % % % % % % % % % % % % % % % % % 
\section{Challenges and Open Problems}
\label{sec:challenge}
The frameworks and principles presented thus far provide a basis for understanding how robustness arises, yet they also make clear that essential scientific and engineering questions remain unanswered. This section identifies some of the open problems and discusses the challenges that must be overcome to achieve manipulation robustness under real-world challenges.

\subsection{Benchmarking and Evaluation of Manipulation Robustness}
\label{challenge:benchmark-eval}
Effective comparison of manipulation robustness across different methods requires well-designed benchmarks and consistent evaluation methods. Simulation-based benchmarking~\citep{tao2024maniskill3,james2020rlbench,zhu2020robosuite} allows experiments under identical conditions and facilitates large-scale testing, yet faces the persistent challenge of the sim-to-real gap that limits physical realism and transferability, as well as being confined to a very narrow distribution of tasks and environments. Real-world benchmarking efforts, including standardized object sets~\citep{calli2017yale}, manipulation competitions~\citep{correll2016analysis,kasper2012kit}, cloud-based robotic platforms~\citep{zahid2024cloudgripper}, and community-run distributed infrastructures~\citep{chen2026manipulationnet,atreya2025roboarena}, aim to address this gap, yet each makes trade-offs among task diversity, scalability, accessibility, authenticity, or consistency. Future progress depends on balancing these aspects through advances in computation, communication, and standardized cross-community protocols.
Most importantly, the existing simulation or real-world benchmarks \emph{rarely} treat robustness as an explicit evaluation criterion, instead evaluating performance primarily under relatively fixed conditions. Fair comparison further requires robustness evaluation methods that quantify system performance. While analytical and empirical methods have been developed for specific manipulation tasks (Section~\ref{sec:evaluation}), a unified framework that systematically captures real-world uncertainty or variation remains an open challenge.

\subsection{Analytical and Empirical Robustness}
\label{challenge:analytical-empirical}
Consistent with the analytical and empirical evaluation protocols discussed in Section~\ref{sec:evaluation}, these two methodological traditions also appear broadly in robustness research. Analytical robustness methods reflect Plato’s philosophy of abstraction, seeking general principles and guarantees through formal modeling and analysis; empirical methods follow Aristotle’s practical spirit, emphasizing performance validated through data and experience. The field continues to debate whether progress in robotics lies in one paradigm or their synthesis~\citep{amato2025data}, and robustness as a core robotic topic follows the same divide. 
Data offer robots experiential knowledge that models alone cannot provide, enabling robustness to emerge from interaction, much as humans develop know-how before know-why through sensorimotor learning. 
Yet unlike human toddlers, robots are not bound to start from scratch. They can inherit models and priors designed by humans, effectively knowing why before knowing how. 
Such models allow robots to adapt behavior in a data-efficient way while mitigating the limitations of purely data-driven methods, which are often sensitive to aleatoric uncertainty and distribution shift. Consider the game of Jenga: humans learn to play robustly not only through trial and error but also by leveraging internal models of perception, planning, and control. Similarly, the future of manipulation robustness likely lies in unifying data-driven learning and model-based reasoning—integrating know-how and know-why into robot systems.

\subsection{Towards Human-Level Manipulation Robustness} 
\label{challenge:toward-human}
Achieving human-level manipulation robustness remains a grand challenge in robotics. Evidence suggests that such robustness emerges from the integration of multiple complementary strategies, such as extensive lifetime experience, accurate world models grounded in physical reasoning, domain-specific embodiment, and rapid adaptation. For example, crows bend hooks to retrieve food, effectively co-designing hardware and control in the wild~\citep{hunt2004crafting}; such behavior cannot be explained by mechanics or control alone. Humans similarly integrate rich lifetime priors with online exploration and adaptation, inspiring approaches such as reinforcement learning bootstrapped with prior data~\citep{luo2025precise}. Yet, synthesizing all these strategies in a single robotic system remains rare and challenging. Other principles evident in animals and humans, such as redundancy in planning and sensing, are gaining attention as essential components for manipulation robustness. Reaching animal-level manipulation robustness already exceeds current robotic capabilities, but is prospective; attaining human-level robustness represents a far greater leap and the ultimate aspiration for robotic manipulation.

\section{Conclusion}
This paper makes advances towards a systematic understanding of manipulation robustness by defining the concept, formulating the problem, and revisiting prior work across subfields. We revisited representative mechanisms and evaluation methods, highlighting the core principles that enable robustness in robotic manipulation and providing insights for future practitioners interested in this topic. Despite decades of progress, achieving human-level manipulation robustness on robotic systems under real-world uncertainties and variations remains a central challenge. 
Bridging this gap will require not only improved algorithms and hardware, but also clearer formulations of robustness and more consistent evaluation methods. We hope this paper will help organize future research efforts toward robotic manipulation systems that routinely approach the robustness demonstrated by humans and animals in the physical world.

{\small 
\section*{Acknowledgements}
The project is partially funded by the European Commission under the Horizon Europe Framework Program project SoftEnable, grant number 101070600.
The authors thank Oliver Brock, Matthew T. Mason, Yan Zhang, Yunke Ao, Rafael I. Cabral Muchacho, Shaohang Han, Haoyu Li, Zizhe Zhang, and Jinda Cui for helpful discussions, comments, or proofreading. Part of Fig.~\ref{fig:overview},~\ref{fig:planning}-c were created using generative AI tools (ChatGPT/OpenAI). These figures are original representative illustrations intended for conceptual explanation. The authors verified their technical accuracy and take full responsibility for the final content.
}

{\small 
\section*{Declaration of conflicting interests} 
The authors declared no potential conflicts of interest with respect to the research,
authorship, and/or publication of this article.
}

\bibliographystyle{SageH}
\bibliography{ref}

@article{madry2018towards,
  title={Towards Deep Learning Models Resistant to Adversarial Attacks},
  author={Madry, Aleksander and Makelov, Aleksandar and Schmidt, Ludwig and Tsipras, Dimitris and Vladu, Adrian},
  journal={International Conference on Learning Representations},
  year={2018}
}

@inproceedings{hendrycks2019benchmarking,
  title={Benchmarking Neural Network Robustness to Common Corruptions and Perturbations},
  author={Hendrycks, Dan and Dietterich, Thomas},
  booktitle={International Conference on Learning Representations},
  year={2019}
}

@book{skogestad2005multivariable,
  title={Multivariable feedback control: analysis and design},
  author={Skogestad, Sigurd and Postlethwaite, Ian},
  year={2005},
  publisher={john Wiley \& sons}
}

@article{moos2022robust,
  title={Robust reinforcement learning: A review of foundations and recent advances},
  author={Moos, Janosch and Hansel, Kay and Abdulsamad, Hany and Stark, Svenja and Clever, Debora and Peters, Jan},
  journal={Machine Learning and Knowledge Extraction},
  volume={4},
  number={1},
  pages={276--315},
  year={2022},
  publisher={MDPI}
}

@inproceedings{ghazi2017morphological,
  title={Morphological computation: the good, the bad, and the ugly},
  author={Ghazi-Zahedi, Keyan and Deimel, Raphael and Mont{\'u}far, Guido and Wall, Vincent and Brock, Oliver},
  booktitle={2017 IEEE/RSJ International Conference on Intelligent Robots and Systems (IROS)},
  pages={464--469},
  year={2017},
  organization={IEEE}
}

@article{cao2022real,
  title={Real-time, highly accurate robotic grasp detection utilizing transfer learning for robots manipulating fragile fruits with widely variable sizes and shapes},
  author={Cao, Boyuan and Zhang, Baohua and Zheng, Wei and Zhou, Jun and Lin, Yihuan and Chen, Yuxin},
  journal={Computers and electronics in agriculture},
  volume={200},
  pages={107254},
  year={2022},
  publisher={Elsevier}
}

@misc{onlinedelivery_mixed_fruit_basket_2kg,
  author       = {{OnlineDelivery.in}},
  title        = {Mixed Fruits with Basket (2 kg)},
  year         = {2024},
  howpublished = {\url{https://www.onlinedelivery.in/mixed-fruits-with-basket-2-kg}},
  note         = {Accessed: 2026-03-13}
}

@article{pokorny2015data,
  title={Data-driven topological motion planning with persistent cohomology},
  author={Pokorny, Florian T and Kragic, Danica},
  journal={2015 Robotics: Science and Systems Conference},
  volume={11},
  year={2015},
  organization={MIT Press}
}

@misc{bair_dex_manip_image1,
  author       = {{Berkeley AI Research}},
  title        = {Dexterous Manipulation Blog Image},
  year         = {2019},
  howpublished = {\url{https://bair.berkeley.edu/static/blog/dex_manip/missing_img1.png}},
  note         = {Image from BAIR blog on dexterous manipulation. Accessed: 2026-03-13}
}

@misc{fruithunters_banana_variety_box,
  author       = {{Fruit Hunters}},
  title        = {Banana Variety Box},
  year         = {2026},
  howpublished = {\url{https://fruithunters.com/products/banana-variety-box}},
  note         = {Accessed: 2026-03-13}
}

@article{sankar2025natural,
  title={A natural biomimetic prosthetic hand with neuromorphic tactile sensing for precise and compliant grasping},
  author={Sankar, Sriramana and Cheng, Wen-Yu and Zhang, Jinghua and Slepyan, Ariel and Iskarous, Mark M and Greene, Rebecca J and DeBrabander, Rene and Chen, Junjun and Gupta, Arnav and Thakor, Nitish V},
  journal={Science Advances},
  volume={11},
  number={10},
  pages={eadr9300},
  year={2025},
  publisher={American Association for the Advancement of Science}
}

@article{lu2025grasping,
  title={Grasping a Handful: Sequential Multi-Object Dexterous Grasp Generation},
  author={Lu, Haofei and Dong, Yifei and Weng, Zehang and Pokorny, Florian and Lundell, Jens and Kragic, Danica},
  journal={IEEE Robotics and Automation Letters},
  year={2025},
  publisher={IEEE}
}

@article{foster1982captive,
  title={A captive behavioral enrichment study with Asian small-clawed river otters (Aonyx cinerea)},
  author={Foster-Turley, Pat and Markowitz, Hal},
  journal={Zoo biology},
  volume={1},
  number={1},
  pages={29--43},
  year={1982},
  publisher={Wiley Online Library}
}

@inproceedings{dafle2014extrinsic,
  title={Extrinsic dexterity: In-hand manipulation with external forces},
  author={Dafle, Nikhil Chavan and Rodriguez, Alberto and Paolini, Robert and Tang, Bowei and Srinivasa, Siddhartha S and Erdmann, Michael and Mason, Matthew T and Lundberg, Ivan and Staab, Harald and Fuhlbrigge, Thomas},
  booktitle={2014 IEEE International Conference on Robotics and Automation (ICRA)},
  pages={1578--1585},
  year={2014},
  organization={IEEE}
}

@article{fenchel1980suspension,
  title={Suspension feeding in ciliated protozoa: functional response and particle size selection},
  author={Fenchel, Tom},
  journal={Microbial Ecology},
  volume={6},
  number={1},
  pages={1--11},
  year={1980},
  publisher={Springer}
}

@article{hadjiosif2015flexible,
  title={Flexible control of safety margins for action based on environmental variability},
  author={Hadjiosif, Alkis M and Smith, Maurice A},
  journal={Journal of Neuroscience},
  volume={35},
  number={24},
  pages={9106--9121},
  year={2015},
  publisher={Society for Neuroscience}
}

@article{cui2021toward,
  title={Toward next-generation learned robot manipulation},
  author={Cui, Jinda and Trinkle, Jeff},
  journal={Science robotics},
  volume={6},
  number={54},
  pages={eabd9461},
  year={2021},
  publisher={American Association for the Advancement of Science}
}

@article{bhattacharya2015persistent,
  title={Persistent homology for path planning in uncertain environments},
  author={Bhattacharya, Subhrajit and Ghrist, Robert and Kumar, Vijay},
  journal={IEEE Transactions on Robotics},
  volume={31},
  number={3},
  pages={578--590},
  year={2015},
  publisher={IEEE}
}

@article{mason2012creation,
  title={Creation myths: The beginnings of robotics research},
  author={Mason, Matthew T},
  journal={IEEE robotics \& automation magazine},
  volume={19},
  number={2},
  pages={72--77},
  year={2012},
  publisher={IEEE}
}

@incollection{braiek2025machine,
  title={Machine learning robustness: A primer},
  author={Braiek, Houssem Ben and Khomh, Foutse},
  booktitle={Trustworthy AI in Medical Imaging},
  pages={37--71},
  year={2025},
  publisher={Elsevier}
}

@article{mason2018toward,
  title={Toward robotic manipulation},
  author={Mason, Matthew T},
  journal={Annual Review of Control, Robotics, and Autonomous Systems},
  volume={1},
  number={1},
  pages={1--28},
  year={2018},
  publisher={Annual Reviews}
}

@book{baum2024robustness,
  title={Robustness in robotic and biological manipulation},
  author={Baum, Manuel},
  year={2024},
  publisher={Technische Universitaet Berlin (Germany)}
}

@article{brunke2022safe,
  title={Safe learning in robotics: From learning-based control to safe reinforcement learning},
  author={Brunke, Lukas and Greeff, Melissa and Hall, Adam W and Yuan, Zhaocong and Zhou, Siqi and Panerati, Jacopo and Schoellig, Angela P},
  journal={Annual Review of Control, Robotics, and Autonomous Systems},
  volume={5},
  number={1},
  pages={411--444},
  year={2022},
  publisher={Annual Reviews}
}

@article{goertz1952fundamentals,
  title={Fundamentals of general-purpose remote manipulators},
  author={Goertz, Raymond C},
  journal={Nucleonics},
  pages={36--42},
  year={1952}
}

@article{billard2019trends,
  title={Trends and challenges in robot manipulation},
  author={Billard, Aude and Kragic, Danica},
  journal={Science},
  volume={364},
  number={6446},
  pages={eaat8414},
  year={2019},
  publisher={American Association for the Advancement of Science}
}

@article{jiang2025robust,
  title={Robust model-based in-hand manipulation with integrated real-time motion-contact planning and tracking},
  author={Jiang, Yongpeng and Yu, Mingrui and Zhu, Xinghao and Tomizuka, Masayoshi and Li, Xiang},
  journal={arXiv preprint arXiv:2505.04978},
  year={2025}
}

@article{xue2025reactive,
  title={Reactive diffusion policy: Slow-fast visual-tactile policy learning for contact-rich manipulation},
  author={Xue, Han and Ren, Jieji and Chen, Wendi and Zhang, Gu and Fang, Yuan and Gu, Guoying and Xu, Huazhe and Lu, Cewu},
  journal={Robotics: Science and Systems},
  year={2025}
}

@inproceedings{dimeas2015reinforcement,
  title={Reinforcement learning of variable admittance control for human-robot co-manipulation},
  author={Dimeas, Fotios and Aspragathos, Nikos},
  booktitle={2015 IEEE/RSJ International Conference on Intelligent Robots and Systems (IROS)},
  pages={1011--1016},
  year={2015},
  organization={IEEE}
}

@article{suh2025dexterous,
  title={Dexterous contact-rich manipulation via the contact trust region},
  author={Suh, HJ Terry and Pang, Tao and Zhao, Tong and Tedrake, Russ},
  journal={The International Journal of Robotics Research},
  pages={02783649251398875},
  year={2025},
  publisher={SAGE Publications Sage UK: London, England}
}

@inproceedings{hou2025adaptive,
  title={Adaptive compliance policy: Learning approximate compliance for diffusion guided control},
  author={Hou, Yifan and Liu, Zeyi and Chi, Cheng and Cousineau, Eric and Kuppuswamy, Naveen and Feng, Siyuan and Burchfiel, Benjamin and Song, Shuran},
  booktitle={2025 IEEE International Conference on Robotics and Automation (ICRA)},
  pages={4829--4836},
  year={2025},
  organization={IEEE}
}

@inproceedings{kamijo2024learning,
  title={Learning variable compliance control from a few demonstrations for bimanual robot with haptic feedback teleoperation system},
  author={Kamijo, Tatsuya and Beltran-Hernandez, Cristian C and Hamaya, Masashi},
  booktitle={2024 IEEE/RSJ International Conference on Intelligent Robots and Systems (IROS)},
  pages={12663--12670},
  year={2024},
  organization={IEEE}
}

@article{xu2026compliant,
  title={Compliant Residual DAgger: Improving Real-World Contact-Rich Manipulation with Human Corrections},
  author={Xu, Xiaomeng and Hou, Yifan and Liu, Zeyi and Song, Shuran},
  journal={The Thirty-ninth Annual Conference on Neural Information Processing Systems},
  year={2026},
}

@article{hogan1984impedance,
  title={Impedance control of industrial robots},
  author={Hogan, Neville},
  journal={Robotics and computer-integrated manufacturing},
  volume={1},
  number={1},
  pages={97--113},
  year={1984},
  publisher={Elsevier}
}

@article{mason2007compliance,
  title={Compliance and force control for computer controlled manipulators},
  author={Mason, Matthew T},
  journal={IEEE Transactions on Systems, Man, and Cybernetics},
  volume={11},
  number={6},
  pages={418--432},
  year={2007},
  publisher={IEEE}
}

@article{holmes2020reachable,
  title={Reachable sets for safe, real-time manipulator trajectory design},
  author={Holmes, Patrick and Kousik, Shreyas and Zhang, Bohao and Raz, Daphna and Barbalata, Corina and Johnson-Roberson, Matthew and Vasudevan, Ram},
  journal={Robotics: Science and Systems},
  year={2020}
}

@article{polverini2017implicit,
  title={Implicit robot force control based on set invariance},
  author={Polverini, Matteo Parigi and Nicolis, Davide and Zanchettin, Andrea Maria and Rocco, Paolo},
  journal={IEEE Robotics and Automation Letters},
  volume={2},
  number={3},
  pages={1288--1295},
  year={2017},
  publisher={IEEE}
}

@article{wang2025caging,
  title={Caging in time: A framework for robust object manipulation under uncertainties and limited robot perception},
  author={Wang, Gaotian and Ren, Kejia and Morgan, Andrew S and Hang, Kaiyu},
  journal={The International Journal of Robotics Research},
  pages={02783649251343926},
  year={2025},
  publisher={SAGE Publications Sage UK: London, England}
}

@article{nubert2020safe,
  title={Safe and fast tracking on a robot manipulator: Robust mpc and neural network control},
  author={Nubert, Julian and K{\"o}hler, Johannes and Berenz, Vincent and Allg{\"o}wer, Frank and Trimpe, Sebastian},
  journal={IEEE Robotics and Automation Letters},
  volume={5},
  number={2},
  pages={3050--3057},
  year={2020},
  publisher={IEEE}
}

@article{flanagan2006control,
  title={Control strategies in object manipulation tasks},
  author={Flanagan, J Randall and Bowman, Miles C and Johansson, Roland S},
  journal={Current opinion in neurobiology},
  volume={16},
  number={6},
  pages={650--659},
  year={2006},
  publisher={Elsevier}
}

@article{koval2016pre,
  title={Pre-and post-contact policy decomposition for planar contact manipulation under uncertainty},
  author={Koval, Michael C and Pollard, Nancy S and Srinivasa, Siddhartha S},
  journal={The International Journal of Robotics Research},
  volume={35},
  number={1-3},
  pages={244--264},
  year={2016},
  publisher={SAGE Publications Sage UK: London, England}
}

@article{lozano1984automatic,
  title={Automatic synthesis of fine-motion strategies for robots},
  author={Lozano-Perez, Tomas and Mason, Matthew T and Taylor, Russell H},
  journal={The International Journal of Robotics Research},
  volume={3},
  number={1},
  pages={3--24},
  year={1984},
  publisher={Sage Publications Sage CA: Thousand Oaks, CA}
}

@article{tedrake2009lqr,
  title={LQR-trees: Feedback motion planning on sparse randomized trees},
  author={Tedrake, R},
  journal={Robotics: Science and Systems V},
  year={2009},
  publisher={Robotics: Science and Systems Foundation}
}

@inproceedings{fox2006exploration,
  title={Exploration of the robustness of plans},
  author={Fox, Maria and Howey, Richard and Long, Derek},
  booktitle={AAAI},
  pages={834--839},
  year={2006}
}

@article{platt2010belief,
  title={Belief space planning assuming maximum likelihood observations},
  author={Platt, R and Tedrake, R and Kaelbling, L and Lozano-Perez, T},
  journal={Robotics: Science and Systems VI},
  year={2010},
  publisher={Robotics: Science and Systems Foundation}
}

@article{kaelbling2013integrated,
  title={Integrated task and motion planning in belief space},
  author={Kaelbling, Leslie Pack and Lozano-P{\'e}rez, Tom{\'a}s},
  journal={The International Journal of Robotics Research},
  volume={32},
  number={9-10},
  pages={1194--1227},
  year={2013},
  publisher={Sage Publications Sage UK: London, England}
}

@article{jankowski2024robust,
  title={Robust pushing: Exploiting quasi-static belief dynamics and contact-informed optimization},
  author={Jankowski, Julius and Bruderm{\"u}ller, Lara and Hawes, Nick and Calinon, Sylvain},
  journal={The International Journal of Robotics Research},
  pages={02783649251318046},
  year={2024},
  publisher={SAGE Publications Sage UK: London, England}
}

@article{bhatt2021surprisingly,
  title={Surprisingly Robust In-Hand Manipulation: An Empirical Study},
  author={Bhatt, Aditya and Sieler, Adrian and Puhlmann, Steffen and Brock, Oliver},
  journal={Robotics: Science and Systems XVII},
  year={2021},
  publisher={Robotics: Science and Systems Foundation}
}

@article{rodriguez2021unstable,
  title={The unstable queen: Uncertainty, mechanics, and tactile feedback},
  author={Rodriguez, Alberto},
  journal={Science Robotics},
  volume={6},
  number={54},
  pages={eabi4667},
  year={2021},
  publisher={American Association for the Advancement of Science}
}

@inproceedings{mason1985mechanics,
  title={The mechanics of manipulation},
  author={Mason, Matthew},
  booktitle={Proceedings. 1985 IEEE International Conference on Robotics and Automation},
  volume={2},
  pages={544--548},
  year={1985},
  organization={IEEE}
}

@article{peshkin2002planning,
  title={Planning robotic manipulation strategies for workpieces that slide},
  author={Peshkin, Michael A and Sanderson, Arthur C},
  journal={IEEE Journal on Robotics and Automation},
  volume={4},
  number={5},
  pages={524--531},
  year={2002},
  publisher={IEEE}
}

@article{dogar2012physics,
  title={Physics-based grasp planning through clutter},
  author={Dogar, M and Hsiao, Kaijen and Ciocarlie, Matei and Srinivasa, Siddhartha},
  journal={Robotics: Science and System},
  pages={57--64},
  year={2012}
}

@article{erdmann2002exploration,
  title={An exploration of sensorless manipulation},
  author={Erdmann, Michael A and Mason, Matthew T},
  journal={IEEE Journal on Robotics and Automation},
  volume={4},
  number={4},
  pages={369--379},
  year={2002},
  publisher={IEEE}
}

@article{blackmore2011chance,
  title={Chance-constrained optimal path planning with obstacles},
  author={Blackmore, Lars and Ono, Masahiro and Williams, Brian C},
  journal={IEEE Transactions on Robotics},
  volume={27},
  number={6},
  pages={1080--1094},
  year={2011},
  publisher={IEEE}
}

@article{shirai2022chance,
  title={Chance-constrained optimization in contact-rich systems for robust manipulation},
  author={Shirai, Yuki and Jha, Devesh K and Raghunathan, Arvind and Romeres, Diego},
  journal={arXiv preprint arXiv:2203.02616},
  year={2022}
}

@article{mayne2005robust,
  title={Robust model predictive control of constrained linear systems with bounded disturbances},
  author={Mayne, David Q and Seron, Mar{\'\i}a M and Rakovi{\'c}, Sa{\v{s}}a V},
  journal={Automatica},
  volume={41},
  number={2},
  pages={219--224},
  year={2005},
  publisher={Elsevier}
}

@inproceedings{berenson2009manipulation,
  title={Manipulation planning on constraint manifolds},
  author={Berenson, Dmitry and Srinivasa, Siddhartha S and Ferguson, Dave and Kuffner, James J},
  booktitle={2009 IEEE international conference on robotics and automation},
  pages={625--632},
  year={2009},
  organization={IEEE}
}

@article{simeon2004manipulation,
  title={Manipulation planning with probabilistic roadmaps},
  author={Sim{\'e}on, Thierry and Laumond, Jean-Paul and Cort{\'e}s, Juan and Sahbani, Anis},
  journal={The International Journal of Robotics Research},
  volume={23},
  number={7-8},
  pages={729--746},
  year={2004},
  publisher={SAGE Publications}
}

@article{saha2007manipulation,
  title={Manipulation planning for deformable linear objects},
  author={Saha, Mitul and Isto, Pekka},
  journal={IEEE Transactions on Robotics},
  volume={23},
  number={6},
  pages={1141--1150},
  year={2007},
  publisher={IEEE}
}

@article{pereira2004decentralized,
  title={Decentralized algorithms for multi-robot manipulation via caging},
  author={Pereira, Guilherme AS and Campos, Mario FM and Kumar, Vijay},
  journal={The International Journal of Robotics Research},
  volume={23},
  number={7-8},
  pages={783--795},
  year={2004},
  publisher={SAGE Publications}
}

@article{howard1996stability,
  title={On the stability of grasped objects},
  author={Howard, W Stamps and Kumar, Vijay},
  journal={IEEE transactions on robotics and automation},
  volume={12},
  number={6},
  pages={904--917},
  year={1996},
  publisher={IEEE}
}

@article{bicchi1995closure,
  title={On the closure properties of robotic grasping},
  author={Bicchi, Antonio},
  journal={The International Journal of Robotics Research},
  volume={14},
  number={4},
  pages={319--334},
  year={1995},
  publisher={Sage Publications Sage CA: Thousand Oaks, CA}
}

@article{aloimonos1988active,
  title={Active vision},
  author={Aloimonos, John and Weiss, Isaac and Bandyopadhyay, Amit},
  journal={International journal of computer vision},
  volume={1},
  number={4},
  pages={333--356},
  year={1988},
  publisher={Springer}
}

@article{xu2015autoscanning,
  title={Autoscanning for coupled scene reconstruction and proactive object analysis},
  author={Xu, Kai and Huang, Hui and Shi, Yifei and Li, Hao and Long, Pinxin and Caichen, Jianong and Sun, Wei and Chen, Baoquan},
  journal={ACM Transactions on Graphics (TOG)},
  volume={34},
  number={6},
  pages={1--14},
  year={2015},
  publisher={ACM New York, NY, USA}
}

@article{bajcsy1988active,
  title={Active perception},
  author={Bajcsy, Ruzena},
  journal={Proceedings of the IEEE},
  volume={76},
  number={8},
  pages={966--1005},
  year={1988},
  publisher={IEEE}
}

@article{chappell2012build,
  title={How to build an information gathering and processing system: Lessons from naturally and artificially intelligent systems},
  author={Chappell, Jackie and Demery, Zoe P and Arriola-Rios, Veronica and Sloman, Aaron},
  journal={Behavioural Processes},
  volume={89},
  number={2},
  pages={179--186},
  year={2012},
  publisher={Elsevier}
}

@article{bohg2017interactive,
  title={Interactive perception: Leveraging action in perception and perception in action},
  author={Bohg, Jeannette and Hausman, Karol and Sankaran, Bharath and Brock, Oliver and Kragic, Danica and Schaal, Stefan and Sukhatme, Gaurav S},
  journal={IEEE Transactions on Robotics},
  volume={33},
  number={6},
  pages={1273--1291},
  year={2017},
  publisher={IEEE}
}

@inproceedings{morrison2019multi,
  title={Multi-view picking: Next-best-view reaching for improved grasping in clutter},
  author={Morrison, Douglas and Corke, Peter and Leitner, J{\"u}rgen},
  booktitle={2019 International Conference on Robotics and Automation (ICRA)},
  pages={8762--8768},
  year={2019},
  organization={IEEE}
}

@inproceedings{li2023see,
  title={See, Hear, and Feel: Smart Sensory Fusion for Robotic Manipulation},
  author={Li, Hao and Zhang, Yizhi and Zhu, Junzhe and Wang, Shaoxiong and Lee, Michelle A and Xu, Huazhe and Adelson, Edward and Fei-Fei, Li and Gao, Ruohan and Wu, Jiajun},
  booktitle={Conference on Robot Learning},
  pages={1368--1378},
  year={2023},
  organization={PMLR}
}

@inproceedings{izatt2017tracking,
  title={Tracking objects with point clouds from vision and touch},
  author={Izatt, Gregory and Mirano, Geronimo and Adelson, Edward and Tedrake, Russ},
  booktitle={2017 IEEE International Conference on Robotics and Automation (ICRA)},
  pages={4000--4007},
  year={2017},
  organization={IEEE}
}

@article{homberg2019robust,
  title={Robust proprioceptive grasping with a soft robot hand},
  author={Homberg, Bianca S and Katzschmann, Robert K and Dogar, Mehmet R and Rus, Daniela},
  journal={Autonomous robots},
  volume={43},
  number={3},
  pages={681--696},
  year={2019},
  publisher={Springer}
}

@book{martin2018leveraging,
  title={Leveraging problem structure in interactive perception for robot manipulation of constrained mechanisms},
  author={Mart{\'\i}n-Mart{\'\i}n, Roberto},
  year={2018},
  publisher={Technische Universitaet Berlin (Germany)}
}

@inproceedings{katz2008manipulating,
  title={Manipulating articulated objects with interactive perception},
  author={Katz, Dov and Brock, Oliver},
  booktitle={2008 IEEE International Conference on Robotics and Automation},
  pages={272--277},
  year={2008},
  organization={IEEE}
}

@article{oquab2023dinov2,
  title={DINOv2: Learning Robust Visual Features without Supervision},
  author={Oquab, Maxime and Darcet, Timoth{'e}e and Moutakanni, Th{'e}o and Vo, Huy and Szafraniec, Marc and Khalidov, Vasil and Fernandez, Pierre and Haziza, Daniel and Massa, Francisco and El-Nouby, Alaaeldin and Assran, Mahmoud and Ballas, Nicolas and Galuba, Wojciech and Howes, Russell and Huang, Po-Yao and Li, Shang-Wen and Misra, Ishan and Rabbat, Michael and Sharma, Vasu and Synnaeve, Gabriel and Xu, Hu and J{'e}gou, Herv{'e} and Mairal, Julien and Labatut, Patrick and Joulin, Armand and Bojanowski, Piotr},
  journal={Transactions on Machine Learning Research Journal},
  year={2024}
}

@article{krizhevsky2012imagenet,
  title={Imagenet classification with deep convolutional neural networks},
  author={Krizhevsky, Alex and Sutskever, Ilya and Hinton, Geoffrey E},
  journal={Advances in neural information processing systems},
  volume={25},
  year={2012}
}

@article{lowe2004distinctive,
  title={Distinctive image features from scale-invariant keypoints},
  author={Lowe, David G},
  journal={International journal of computer vision},
  volume={60},
  number={2},
  pages={91--110},
  year={2004},
  publisher={Springer}
}

@inproceedings{bay2006surf,
  title={Surf: Speeded up robust features},
  author={Bay, Herbert and Tuytelaars, Tinne and Van Gool, Luc},
  booktitle={European conference on computer vision},
  pages={404--417},
  year={2006},
  organization={Springer}
}

@article{lecun2002gradient,
  title={Gradient-based learning applied to document recognition},
  author={LeCun, Yann and Bottou, L{\'e}on and Bengio, Yoshua and Haffner, Patrick},
  journal={Proceedings of the IEEE},
  volume={86},
  number={11},
  pages={2278--2324},
  year={2002},
  publisher={Ieee}
}

@inproceedings{zitkovich2023rt,
  title={Rt-2: Vision-language-action models transfer web knowledge to robotic control},
  author={Zitkovich, Brianna and Yu, Tianhe and Xu, Sichun and Xu, Peng and Xiao, Ted and Xia, Fei and Wu, Jialin and Wohlhart, Paul and Welker, Stefan and Wahid, Ayzaan and Vuong, Quan and Vanhoucke, Vincent and Tran, Huong and Soricut, Radu and Singh, Anikait and Singh, Jaspiar and Sermanet, Pierre and Sanketi, Pannag R. and Salazar, Grecia and Ryoo, Michael S. and Reymann, Krista and Rao, Kanishka and Pertsch, Karl and Mordatch, Igor and Michalewski, Henryk and Lu, Yao and Levine, Sergey and Lee, Lisa and Lee, Tsang-Wei Edward and Leal, Isabel and Kuang, Yuheng and Kalashnikov, Dmitry and Julian, Ryan and Joshi, Nikhil J. and Irpan, Alex and Ichter, Brian and Hsu, Jasmine and Herzog, Alexander and Hausman, Karol and Gopalakrishnan, Keerthana and Fu, Chuyuan and Florence, Pete and Finn, Chelsea and Dubey, Kumar Avinava and Driess, Danny and Ding, Tianli and Choromanski, Krzysztof Marcin and Chen, Xi and Chebotar, Yevgen and Carbajal, Justice and Brown, Noah and Brohan, Anthony and Arenas, Montserrat Gonzalez and Han, Kehang},
  booktitle={Conference on Robot Learning},
  pages={2165--2183},
  year={2023},
  organization={PMLR}
}

@inproceedings{rublee2011orb,
  title={ORB: An efficient alternative to SIFT or SURF},
  author={Rublee, Ethan and Rabaud, Vincent and Konolige, Kurt and Bradski, Gary},
  booktitle={2011 International conference on computer vision},
  pages={2564--2571},
  year={2011},
  organization={Ieee}
}

@misc{autumn2006tokayfoot,
  author       = {Autumn, Kellar},
  title        = {Tokay gecko foot image},
  year         = {2006},
  howpublished = {\url{https://people.eecs.berkeley.edu/~ronf/Gecko/Interface-slide-adhesion/TokayFoot2-KA.jpg}},
  note         = {Copyright \textcopyright\ 2006 Kellar Autumn. Accessed: 2026-06-21}
}

@article{paul2006morphological,
  title={Morphological computation: A basis for the analysis of morphology and control requirements},
  author={Paul, Chandana},
  journal={Robotics and Autonomous Systems},
  volume={54},
  number={8},
  pages={619--630},
  year={2006},
  publisher={Elsevier}
}

@article{mann2008dolphins,
  title={Why do dolphins carry sponges?},
  author={Mann, Janet and Sargeant, Brooke L and Watson-Capps, Jana J and Gibson, Quincy A and Heithaus, Michael R and Connor, Richard C and Patterson, Eric},
  journal={PloS one},
  volume={3},
  number={12},
  pages={e3868},
  year={2008},
  publisher={Public Library of Science San Francisco, USA}
}

@phdthesis{drake1978using,
  title={Using compliance in lieu of sensory feedback for automatic assembly.},
  author={Drake, Samuel Hunt},
  year={1978},
  school={Massachusetts Institute of Technology}
}

@inproceedings{deimel2013compliant,
  title={A compliant hand based on a novel pneumatic actuator},
  author={Deimel, Raphael and Brock, Oliver},
  booktitle={2013 IEEE international conference on robotics and automation},
  pages={2047--2053},
  year={2013},
  organization={IEEE}
}

@article{shintake2018soft,
  title={Soft robotic grippers},
  author={Shintake, Jun and Cacucciolo, Vito and Floreano, Dario and Shea, Herbert},
  journal={Advanced materials},
  volume={30},
  number={29},
  pages={1707035},
  year={2018},
  publisher={Wiley Online Library}
}

@article{amend2012positive,
  title={A positive pressure universal gripper based on the jamming of granular material},
  author={Amend, John R and Brown, Eric and Rodenberg, Nicholas and Jaeger, Heinrich M and Lipson, Hod},
  journal={IEEE transactions on robotics},
  volume={28},
  number={2},
  pages={341--350},
  year={2012},
  publisher={IEEE}
}

@article{jiang2017robotic,
  title={A robotic device using gecko-inspired adhesives can grasp and manipulate large objects in microgravity},
  author={Jiang, Hao and Hawkes, Elliot W. and Fuller, Christine and Estrada, Max A. and Suresh, Sai A. and Abcouwer, Nathan and Han, Angela K. and Wang, Shimeng and Ploch, Carly J. and Parness, Aaron and Cutkosky, Mark R.},
  journal={Science Robotics},
  volume={2},
  number={7},
  pages={eaan4545},
  year={2017},
  publisher={American Association for the Advancement of Science}
}

@article{zeng2022robotic,
  title={Robotic pick-and-place of novel objects in clutter with multi-affordance grasping and cross-domain image matching},
  author={Zeng, Andy and Song, Shuran and Yu, Kuan-Ting and Donlon, Elliott and Hogan, Francois R. and Bauza, Maria and Ma, Daolin and Taylor, Orion and Liu, Melody and Romo, Eudald and Fazeli, Nima and Alet, Ferran and Dafle, Nikhil Chavan and Holladay, Rachel and Morona, Isabella and Nair, Prem Qu and Green, Druck and Taylor, Ian and Liu, Weber and Funkhouser, Thomas and Rodriguez, Alberto},
  journal={The International Journal of Robotics Research},
  volume={41},
  number={7},
  pages={690--705},
  year={2022},
  publisher={SAGE Publications Sage UK: London, England}
}

@article{liang2026modular,
  title={Modular reconfigurable robots: Toward on-demand multifunctional applications},
  author={Liang, Guanqi and Ijspeert, Auke Jan and Yim, Mark and Lam, Tin Lun},
  journal={Science Robotics},
  volume={11},
  number={111},
  pages={eadz1999},
  year={2026},
  publisher={American Association for the Advancement of Science}
}

@article{pagoli2021soft,
  title={A soft robotic gripper with an active palm and reconfigurable fingers for fully dexterous in-hand manipulation},
  author={Pagoli, Amir and Chapelle, Fr{\'e}d{\'e}ric and Corrales, Juan Antonio and Mezouar, Youcef and Lapusta, Yuri},
  journal={IEEE Robotics and Automation Letters},
  volume={6},
  number={4},
  pages={7706--7713},
  year={2021},
  publisher={IEEE}
}

@inproceedings{yim2000polybot,
  title={PolyBot: a modular reconfigurable robot},
  author={Yim, Mark and Duff, David G and Roufas, Kimon D},
  booktitle={Proceedings 2000 ICRA. millennium conference. IEEE international conference on robotics and automation. Symposia proceedings (Cat. No. 00CH37065)},
  volume={1},
  pages={514--520},
  year={2000},
  organization={IEEE}
}

@article{xie2019reconfigurable,
  title={Reconfigurable magnetic microrobot swarm: Multimode transformation, locomotion, and manipulation},
  author={Xie, Hui and Sun, Mengmeng and Fan, Xinjian and Lin, Zhihua and Chen, Weinan and Wang, Lei and Dong, Lixin and He, Qiang},
  journal={Science robotics},
  volume={4},
  number={28},
  pages={eaav8006},
  year={2019},
  publisher={American Association for the Advancement of Science}
}

@article{chi2023diffusion,
  title={Diffusion policy: Visuomotor policy learning via action diffusion},
  author={Chi, Cheng and Xu, Zhenjia and Feng, Siyuan and Cousineau, Eric and Du, Yilun and Burchfiel, Benjamin and Tedrake, Russ and Song, Shuran},
  journal={The International Journal of Robotics Research},
  pages={02783649241273668},
  year={2023},
  publisher={SAGE Publications Sage UK: London, England}
}

@inproceedings{tobin2017domain,
  title={Domain randomization for transferring deep neural networks from simulation to the real world},
  author={Tobin, Josh and Fong, Rachel and Ray, Alex and Schneider, Jonas and Zaremba, Wojciech and Abbeel, Pieter},
  booktitle={2017 IEEE/RSJ international conference on intelligent robots and systems (IROS)},
  pages={23--30},
  year={2017},
  organization={IEEE}
}

@article{akkaya2019solving,
title={Solving rubik's cube with a robot hand},
author={{OpenAI} and Akkaya, Ilge and Andrychowicz, Marcin and Chociej, Maciek and Litwin, Mateusz and McGrew, Bob and Petron, Arthur and Paino, Alex and Plappert, Matthias and Powell, Glenn and Ribas, Raphael and Schneider, Jonas and Tezak, Nikolas and Tworek, Jerry and Welinder, Peter and Weng, Lilian and Yuan, Qiming and Zaremba, Wojciech and Zhang, Lei},
journal={arXiv preprint arXiv:1910.07113},
year={2019}
}

@article{andrychowicz2020learning,
  title={Learning dexterous in-hand manipulation},
  author={Andrychowicz, OpenAI: Marcin and Baker, Bowen and Chociej, Maciek and Jozefowicz, Rafal and McGrew, Bob and Pachocki, Jakub and Petron, Arthur and Plappert, Matthias and Powell, Glenn and Ray, Alex and Schneider, Jonas and Sidor, Szymon and Tobin, Josh and Welinder, Peter and Weng, Lilian and Zaremba, Wojciech},
  journal={The International Journal of Robotics Research},
  volume={39},
  number={1},
  pages={3--20},
  year={2020},
  publisher={SAGE Publications Sage UK: London, England}
}

@article{florence2019self,
  title={Self-supervised correspondence in visuomotor policy learning},
  author={Florence, Peter and Manuelli, Lucas and Tedrake, Russ},
  journal={IEEE Robotics and Automation Letters},
  volume={5},
  number={2},
  pages={492--499},
  year={2019},
  publisher={IEEE}
}

@inproceedings{mandlekar2023mimicgen,
  title={MimicGen: A Data Generation System for Scalable Robot Learning using Human Demonstrations},
  author={Mandlekar, Ajay and Nasiriany, Soroush and Wen, Bowen and Akinola, Iretiayo and Narang, Yashraj and Fan, Linxi and Zhu, Yuke and Fox, Dieter},
  booktitle={Conference on Robot Learning},
  pages={1820--1864},
  year={2023},
  organization={PMLR}
}

@inproceedings{ameperosa2025rocoda,
  title={Rocoda: Counterfactual data augmentation for data-efficient robot learning from demonstrations},
  author={Ameperosa, Ezra and Collins, Jeremy A and Jain, Mrinal and Garg, Animesh},
  booktitle={2025 IEEE International Conference on Robotics and Automation (ICRA)},
  pages={13250--13256},
  year={2025},
  organization={IEEE}
}

@inproceedings{zhou2023nerf,
  title={Nerf in the palm of your hand: Corrective augmentation for robotics via novel-view synthesis},
  author={Zhou, Allan and Kim, Moo Jin and Wang, Lirui and Florence, Pete and Finn, Chelsea},
  booktitle={Proceedings of the IEEE/CVF Conference on Computer Vision and Pattern Recognition},
  pages={17907--17917},
  year={2023}
}

@inproceedings{zhang2024diffusion,
  title={Diffusion Meets DAgger: Supercharging Eye-in-hand Imitation Learning},
  author={Zhang, Xiaoyu and Chang, Matthew and Kumar, Pranav and Gupta, Saurabh},
  booktitle={Robotics science and systems},
  year={2024},
  organization={Robotics science and systems}
}

@article{yu2023scaling,
  title={Scaling robot learning with semantically imagined experience},
  author={Yu, Tianhe and Xiao, Ted and Stone, Austin and Tompson, Jonathan and Brohan, Anthony and Wang, Su and Singh, Jaspiar and Tan, Clayton and M, Dee and Peralta, Jodilyn and Ichter, Brian and Hausman, Karol and Xia, Fei},
  journal={Robotics: Science and Systems},
  year={2023}
}

@article{chen2023genaug,
  title={Genaug: Retargeting behaviors to unseen situations via generative augmentation},
  author={Chen, Zoey and Kiami, Sho and Gupta, Abhishek and Kumar, Vikash},
  journal={Robotics: Science and Systems},
  year={2023}
}

@article{xue2025demogen,
  title={Demogen: Synthetic demonstration generation for data-efficient visuomotor policy learning},
  author={Xue, Zhengrong and Deng, Shuying and Chen, Zhenyang and Wang, Yixuan and Yuan, Zhecheng and Xu, Huazhe},
  journal={Robotics: Science and Systems},
  year={2025}
}

@inproceedings{laskey2017dart,
  title={Dart: Noise injection for robust imitation learning},
  author={Laskey, Michael and Lee, Jonathan and Fox, Roy and Dragan, Anca and Goldberg, Ken},
  booktitle={Conference on robot learning},
  pages={143--156},
  year={2017},
  organization={PMLR}
}

@inproceedings{ke2021grasping,
  title={Grasping with chopsticks: Combating covariate shift in model-free imitation learning for fine manipulation},
  author={Ke, Liyiming and Wang, Jingqiang and Bhattacharjee, Tapomayukh and Boots, Byron and Srinivasa, Siddhartha},
  booktitle={2021 IEEE international conference on robotics and automation (ICRA)},
  pages={6185--6191},
  year={2021},
  organization={IEEE}
}

@article{simchowitz2025pitfalls,
  title={The pitfalls of imitation learning when actions are continuous},
  author={Simchowitz, Max and Pfrommer, Daniel and Jadbabaie, Ali},
  journal={arXiv preprint arXiv:2503.09722},
  year={2025}
}

@article{brohan2022rt,
  title={RT-1: Robotics Transformer for Real-World Control at Scale},
  author={Brohan, Anthony and Brown, Noah and Carbajal, Justice and Chebotar, Yevgen and Dabis, Joseph and Finn, Chelsea and Gopalakrishnan, Keerthana and Hausman, Karol and Herzog, Alexander and Hsu, Jasmine and Ibarz, Julian and Ichter, Brian and Irpan, Alex and Jackson, Tomas and Jesmonth, Sally and Joshi, Nikhil and Julian, Ryan and Kalashnikov, Dmitry and Kuang, Yuheng and Leal, Isabel and Lee, Kuang-Huei and Levine, Sergey and Lu, Yao and Malla, Utsav and Manjunath, Deeksha and Mordatch, Igor and Nachum, Ofir and Parada, Carolina and Peralta, Jodilyn and Perez, Emily and Pertsch, Karl and Quiambao, Jornell and Rao, Kanishka and Ryoo, Michael S. and Salazar, Grecia and Sanketi, Pannag R. and Sayed, Kevin and Singh, Jaspiar and Sontakke, Sumedh and Stone, Austin and Tan, Clayton and Tran, Huong and Vanhoucke, Vincent and Vega, Steve and Vuong, Quan H. and Xia, Fei and Xiao, Ted and Xu, Peng and Xu, Sichun and Yu, Tianhe and Zitkovich, Brianna},
  journal={Robotics: Science and Systems XIX},
  year={2023},
  publisher={Robotics: Science and Systems Foundation}
}

@inproceedings{o2024open,
  title={Open x-embodiment: Robotic learning datasets and rt-x models: Open x-embodiment collaboration 0},
  author={{Open X-Embodiment Collaboration}},
  booktitle={2024 IEEE International Conference on Robotics and Automation (ICRA)},
  pages={6892--6903},
  year={2024},
  organization={IEEE}
}

@inproceedings{kim2024openvla,
  title={OpenVLA: An Open-Source Vision-Language-Action Model},
  author={Kim, Moo Jin and Pertsch, Karl and Karamcheti, Siddharth and Xiao, Ted and Balakrishna, Ashwin and Nair, Suraj and Rafailov, Rafael and Foster, Ethan P. and Sanketi, Pannag R. and Vuong, Quan and Kollar, Thomas and Burchfiel, Benjamin and Tedrake, Russ and Sadigh, Dorsa and Levine, Sergey and Liang, Percy and Finn, Chelsea},
  booktitle={Conference on Robot Learning},
  pages={2679--2713},
  year={2025},
  organization={PMLR}
}

@inproceedings{manuelli2019kpam,
  title={kpam: Keypoint affordances for category-level robotic manipulation},
  author={Manuelli, Lucas and Gao, Wei and Florence, Peter and Tedrake, Russ},
  booktitle={The International Symposium of Robotics Research},
  pages={132--157},
  year={2019},
  organization={Springer}
}

@inproceedings{qin2020keto,
  title={Keto: Learning keypoint representations for tool manipulation},
  author={Qin, Zengyi and Fang, Kuan and Zhu, Yuke and Fei-Fei, Li and Savarese, Silvio},
  booktitle={2020 IEEE International Conference on Robotics and Automation (ICRA)},
  pages={7278--7285},
  year={2020},
  organization={IEEE}
}

@inproceedings{huang2024rekep,
  title={ReKep: Spatio-Temporal Reasoning of Relational Keypoint Constraints for Robotic Manipulation},
  author={Huang, Wenlong and Wang, Chen and Li, Yunzhu and Zhang, Ruohan and Fei-Fei, Li},
  booktitle={Conference on Robot Learning},
  pages={4573--4602},
  year={2025},
  organization={PMLR}
}

@inproceedings{li2020towards,
  title={Towards practical multi-object manipulation using relational reinforcement learning},
  author={Li, Richard and Jabri, Allan and Darrell, Trevor and Agrawal, Pulkit},
  booktitle={2020 ieee international conference on robotics and automation (icra)},
  pages={4051--4058},
  year={2020},
  organization={IEEE}
}

@article{lin2022efficient,
  title={Efficient and interpretable robot manipulation with graph neural networks},
  author={Lin, Yixin and Wang, Austin S and Undersander, Eric and Rai, Akshara},
  journal={IEEE Robotics and Automation Letters},
  volume={7},
  number={2},
  pages={2740--2747},
  year={2022},
  publisher={IEEE}
}

@inproceedings{huang2022planning,
  title={Planning for multi-object manipulation with graph neural network relational classifiers},
  author={Huang, Yixuan and Conkey, Adam and Hermans, Tucker},
  booktitle={2023 IEEE International Conference on Robotics and Automation (ICRA)},
  pages={1822--1829},
  year={2023},
  organization={IEEE}
}

@inproceedings{simeonov2022neural,
  title={Neural descriptor fields: Se (3)-equivariant object representations for manipulation},
  author={Simeonov, Anthony and Du, Yilun and Tagliasacchi, Andrea and Tenenbaum, Joshua B and Rodriguez, Alberto and Agrawal, Pulkit and Sitzmann, Vincent},
  booktitle={2022 International Conference on Robotics and Automation (ICRA)},
  pages={6394--6400},
  year={2022},
  organization={IEEE}
}

@article{ryu2022equivariant,
  title={Equivariant descriptor fields: Se (3)-equivariant energy-based models for end-to-end visual robotic manipulation learning},
  author={Ryu, Hyunwoo and Lee, Hong-in and Lee, Jeong-Hoon and Choi, Jongeun},
  journal={arXiv preprint arXiv:2206.08321},
  year={2022}
}

@inproceedings{eisner2024deep,
  title={Deep SE (3)-Equivariant Geometric Reasoning for Precise Placement Tasks},
  author={Eisner, Ben and Yang, Yi and Davchev, Todor and Vecerik, Mel and Scholz, Jonathan and Held, David},
  booktitle={The Twelfth International Conference on Learning Representations},
  year={2024}
}

@inproceedings{wang2024equivariant,
  title={Equivariant Diffusion Policy},
  author={Wang, Dian and Hart, Stephen and Surovik, David and Kelestemur, Tarik and Huang, Haojie and Zhao, Haibo and Yeatman, Mark and Wang, Jiuguang and Walters, Robin and Platt, Robert},
  booktitle={Conference on Robot Learning},
  pages={48--69},
  year={2025},
  organization={PMLR}
}

@inproceedings{sutton1998intra,
  title={Intra-Option Learning about Temporally Abstract Actions.},
  author={Sutton, Richard S and Precup, Doina and Singh, Satinder},
  booktitle={ICML},
  volume={98},
  pages={556--564},
  year={1998}
}

@article{driess2021learning,
  title={Learning to solve sequential physical reasoning problems from a scene image},
  author={Driess, Danny and Ha, Jung-Su and Toussaint, Marc},
  journal={The International Journal of Robotics Research},
  volume={40},
  number={12-14},
  pages={1435--1466},
  year={2021},
  publisher={SAGE Publications Sage UK: London, England}
}

@article{zhu2022bottom,
  title={Bottom-up skill discovery from unsegmented demonstrations for long-horizon robot manipulation},
  author={Zhu, Yifeng and Stone, Peter and Zhu, Yuke},
  journal={IEEE Robotics and Automation Letters},
  volume={7},
  number={2},
  pages={4126--4133},
  year={2022},
  publisher={IEEE}
}

@inproceedings{mishra2023generative,
  title={Generative skill chaining: Long-horizon skill planning with diffusion models},
  author={Mishra, Utkarsh Aashu and Xue, Shangjie and Chen, Yongxin and Xu, Danfei},
  booktitle={Conference on Robot Learning},
  pages={2905--2925},
  year={2023},
  organization={PMLR}
}

@article{black2410pi0, 
    AUTHOR    = {Kevin Black AND Noah Brown AND Danny Driess AND Adnan Esmail AND Michael Robert Equi AND Chelsea Finn AND Niccolo Fusai AND Lachy Groom AND Karol Hausman AND Brian Ichter AND Szymon Jakubczak AND Tim Jones AND Liyiming Ke AND Sergey Levine AND Adrian Li-Bell AND Mohith Mothukuri AND Suraj Nair AND Karl Pertsch AND Lucy Xiaoyang Shi AND Laura Smith AND James Tanner AND Quan Vuong AND Anna Walling AND Haohuan Wang AND Ury Zhilinsky}, 
    TITLE     = {{$\pi$0: A Vision-Language-Action Flow Model for General Robot Control}}, 
    journal = {Proceedings of Robotics: Science and Systems}, 
    YEAR      = {2025}, 
}

@article{pan2025much,
  title={Much Ado About Noising: Dispelling the Myths of Generative Robotic Control},
  author={Pan, Chaoyi and Anantharaman, Giri and Huang, Nai-Chieh and Jin, Claire and Pfrommer, Daniel and Yuan, Chenyang and Permenter, Frank and Qu, Guannan and Boffi, Nicholas and Shi, Guanya and Simchowitz, Max},
  journal={International Conference on Learning Representations},
  year={2026}
}

@inproceedings{hafner2019learning,
  title={Learning latent dynamics for planning from pixels},
  author={Hafner, Danijar and Lillicrap, Timothy and Fischer, Ian and Villegas, Ruben and Ha, David and Lee, Honglak and Davidson, James},
  booktitle={International conference on machine learning},
  pages={2555--2565},
  year={2019},
  organization={PMLR}
}

@article{hafner2023mastering,
  title={Mastering diverse domains through world models},
  author={Hafner, Danijar and Pasukonis, Jurgis and Ba, Jimmy and Lillicrap, Timothy},
  journal={arXiv preprint arXiv:2301.04104},
  year={2023}
}

@inproceedings{hansen2022temporal,
  title={Temporal Difference Learning for Model Predictive Control},
  author={Hansen, N and Wang, X and Su, H},
  booktitle={International Conference on Machine Learning, PMLR},
  year={2022}
}

@article{ichter2019robot,
  title={Robot motion planning in learned latent spaces},
  author={Ichter, Brian and Pavone, Marco},
  journal={IEEE Robotics and Automation Letters},
  volume={4},
  number={3},
  pages={2407--2414},
  year={2019},
  publisher={IEEE}
}

@inproceedings{liu2024model,
  title={Model-based runtime monitoring with interactive imitation learning},
  author={Liu, Huihan and Dass, Shivin and Mart{\'\i}n-Mart{\'\i}n, Roberto and Zhu, Yuke},
  booktitle={2024 IEEE International Conference on Robotics and Automation (ICRA)},
  pages={4154--4161},
  year={2024},
  organization={IEEE}
}

@article{nakamura2025generalizing,
  title={Generalizing safety beyond collision-avoidance via latent-space reachability analysis},
  author={Nakamura, Kensuke and Peters, Lasse and Bajcsy, Andrea},
  journal={arXiv preprint arXiv:2502.00935},
  year={2025}
}

@article{sun2025latent,
  title={Latent policy barrier: Learning robust visuomotor policies by staying in-distribution},
  author={Sun, Zhanyi and Song, Shuran},
  journal={Advances in Neural Information Processing Systems},
  volume={38},
  pages={174280--174305},
  year={2026}
}

@inproceedings{pinto2017robust,
  title={Robust adversarial reinforcement learning},
  author={Pinto, Lerrel and Davidson, James and Sukthankar, Rahul and Gupta, Abhinav},
  booktitle={International conference on machine learning},
  pages={2817--2826},
  year={2017},
  organization={PMLR}
}

@inproceedings{pinto2017supervision,
  title={Supervision via competition: Robot adversaries for learning tasks},
  author={Pinto, Lerrel and Davidson, James and Gupta, Abhinav},
  booktitle={2017 IEEE International Conference on Robotics and Automation (ICRA)},
  pages={1601--1608},
  year={2017},
  organization={IEEE}
}

@inproceedings{jian2021adversarial,
  title={Adversarial skill learning for robust manipulation},
  author={Jian, Pingcheng and Yang, Chao and Guo, Di and Liu, Huaping and Sun, Fuchun},
  booktitle={2021 IEEE International Conference on Robotics and Automation (ICRA)},
  pages={2555--2561},
  year={2021},
  organization={IEEE}
}

@inproceedings{achiam2017constrained,
  title={Constrained policy optimization},
  author={Achiam, Joshua and Held, David and Tamar, Aviv and Abbeel, Pieter},
  booktitle={International conference on machine learning},
  pages={22--31},
  year={2017},
  organization={PMLR}
}

@article{dalal2018safe,
  title={Safe exploration in continuous action spaces},
  author={Dalal, Gal and Dvijotham, Krishnamurthy and Vecerik, Matej and Hester, Todd and Paduraru, Cosmin and Tassa, Yuval},
  journal={arXiv preprint arXiv:1801.08757},
  year={2018}
}

@inproceedings{cheng2019end,
  title={End-to-end safe reinforcement learning through barrier functions for safety-critical continuous control tasks},
  author={Cheng, Richard and Orosz, G{\'a}bor and Murray, Richard M and Burdick, Joel W},
  booktitle={Proceedings of the AAAI conference on artificial intelligence},
  volume={33},
  pages={3387--3395},
  year={2019}
}

@article{levine2020offline,
  title={Offline reinforcement learning: Tutorial, review, and perspectives on open problems},
  author={Levine, Sergey and Kumar, Aviral and Tucker, George and Fu, Justin},
  journal={arXiv preprint arXiv:2005.01643},
  year={2020}
}

@article{kumar2020conservative,
  title={Conservative q-learning for offline reinforcement learning},
  author={Kumar, Aviral and Zhou, Aurick and Tucker, George and Levine, Sergey},
  journal={Advances in neural information processing systems},
  volume={33},
  pages={1179--1191},
  year={2020}
}

@inproceedings{kostrikov2021offline,
  title={Offline Reinforcement Learning with Implicit Q-Learning},
  author={Kostrikov, Ilya and Nair, Ashvin and Levine, Sergey},
  booktitle={International Conference on Learning Representations},
  year={2021}
}

@article{fujimoto2021minimalist,
  title={A minimalist approach to offline reinforcement learning},
  author={Fujimoto, Scott and Gu, Shixiang Shane},
  journal={Advances in neural information processing systems},
  volume={34},
  pages={20132--20145},
  year={2021}
}

@inproceedings{zhou2022real,
  title={Real world offline reinforcement learning with realistic data source},
  author={Zhou, Gaoyue and Ke, Liyiming and Srinivasa, Siddhartha and Gupta, Abhinav and Rajeswaran, Aravind and Kumar, Vikash},
  booktitle={2023 IEEE international conference on robotics and automation (ICRA)},
  pages={7176--7183},
  year={2023},
  organization={IEEE}
}

@article{herzog2023deep,
  title={Deep RL at scale: Sorting waste in office buildings with a fleet of mobile manipulators},
  author={Herzog, Alexander and Rao, Kanishka and Hausman, Karol and Lu, Yao and Wohlhart, Paul and Yan, Mengyuan and Lin, Jessica and Arenas, Montserrat Gonzalez and Xiao, Ted and Kappler, Daniel and Ho, Daniel and Rettinghouse, Jarek and Chebotar, Yevgen and Lee, Kuang-Huei and Gopalakrishnan, Keerthana and Julian, Ryan and Li, Adrian and Fu, Chuyuan Kelly and Wei, Bob and Ramesh, Sangeetha and Holden, Khem and Kleiven, Kim and Rendleman, David and Kirmani, Sean and Bingham, Jeff and Weisz, Jon and Xu, Ying and Lu, Wenlong and Bennice, Matthew and Fong, Cody and Do, David and Lam, Jessica and Bai, Yunfei and Holson, Benjie and Quinlan, Michael and Brown, Noah and Kalakrishnan, Mrinal and Ibarz, Julian and Pastor, Peter and Levine, Sergey},
  journal={Robotics: Science and Systems (RSS)},
  year={2023}
}

@inproceedings{haldar2023watch,
  title={Watch and match: Supercharging imitation with regularized optimal transport},
  author={Haldar, Siddhant and Mathur, Vaibhav and Yarats, Denis and Pinto, Lerrel},
  booktitle={Conference on Robot Learning},
  pages={32--43},
  year={2023},
  organization={PMLR}
}

@inproceedings{ross2011reduction,
  title={A reduction of imitation learning and structured prediction to no-regret online learning},
  author={Ross, St{\'e}phane and Gordon, Geoffrey and Bagnell, Drew},
  booktitle={Proceedings of the fourteenth international conference on artificial intelligence and statistics},
  pages={627--635},
  year={2011},
  organization={JMLR Workshop and Conference Proceedings}
}

@inproceedings{kelly2019hg,
  title={Hg-dagger: Interactive imitation learning with human experts},
  author={Kelly, Michael and Sidrane, Chelsea and Driggs-Campbell, Katherine and Kochenderfer, Mykel J},
  booktitle={2019 International Conference on Robotics and Automation (ICRA)},
  pages={8077--8083},
  year={2019},
  organization={IEEE}
}

@article{mandlekar2020human,
  title={Human-in-the-loop imitation learning using remote teleoperation},
  author={Mandlekar, Ajay and Xu, Danfei and Mart{\'\i}n-Mart{\'\i}n, Roberto and Zhu, Yuke and Fei-Fei, Li and Savarese, Silvio},
  journal={arXiv preprint arXiv:2012.06733},
  year={2020}
}

@article{levine2016end,
  title={End-to-end training of deep visuomotor policies},
  author={Levine, Sergey and Finn, Chelsea and Darrell, Trevor and Abbeel, Pieter},
  journal={Journal of Machine Learning Research},
  volume={17},
  number={39},
  pages={1--40},
  year={2016}
}

@article{rajeswaran2017learning,
  title={Learning Complex Dexterous Manipulation with Deep Reinforcement Learning and Demonstrations},
  author={Rajeswaran, Aravind and Kumar, Vikash and Gupta, Abhishek and Vezzani, Giulia and Schulman, John and Todorov, Emanuel and Levine, Sergey},
  journal={Robotics: Science and Systems XIV},
  year={2018},
  publisher={Robotics: Science and Systems Foundation}
}

@inproceedings{johannink2019residual,
  title={Residual reinforcement learning for robot control},
  author={Johannink, Tobias and Bahl, Shikhar and Nair, Ashvin and Luo, Jianlan and Kumar, Avinash and Loskyll, Matthias and Ojea, Juan Aparicio and Solowjow, Eugen and Levine, Sergey},
  booktitle={2019 international conference on robotics and automation (ICRA)},
  pages={6023--6029},
  year={2019},
  organization={IEEE}
}

@inproceedings{xu2022dexterous,
  title={Dexterous manipulation from images: Autonomous real-world rl via substep guidance},
  author={Xu, Kelvin and Hu, Zheyuan and Doshi, Ria and Rovinsky, Aaron and Kumar, Vikash and Gupta, Abhishek and Levine, Sergey},
  booktitle={2023 IEEE International Conference on Robotics and Automation (ICRA)},
  pages={5938--5945},
  year={2023},
  organization={IEEE}
}

@inproceedings{luo2024serl,
  title={Serl: A software suite for sample-efficient robotic reinforcement learning},
  author={Luo, Jianlan and Hu, Zheyuan and Xu, Charles and Tan, You Liang and Berg, Jacob and Sharma, Archit and Schaal, Stefan and Finn, Chelsea and Gupta, Abhishek and Levine, Sergey},
  booktitle={2024 IEEE International Conference on Robotics and Automation (ICRA)},
  pages={16961--16969},
  year={2024},
  organization={IEEE}
}

@article{duan2017one,
  title={One-shot imitation learning},
  author={Duan, Yan and Andrychowicz, Marcin and Stadie, Bradly and Jonathan Ho, OpenAI and Schneider, Jonas and Sutskever, Ilya and Abbeel, Pieter and Zaremba, Wojciech},
  journal={Advances in neural information processing systems},
  volume={30},
  year={2017}
}

@inproceedings{valassakis2022demonstrate,
  title={Demonstrate once, imitate immediately (dome): Learning visual servoing for one-shot imitation learning},
  author={Valassakis, Eugene and Papagiannis, Georgios and Di Palo, Norman and Johns, Edward},
  booktitle={2022 IEEE/RSJ International Conference on Intelligent Robots and Systems (IROS)},
  pages={8614--8621},
  year={2022},
  organization={IEEE}
}

@article{fu2024context,
  title={In-context imitation learning via next-token prediction},
  author={Fu, Letian and Huang, Huang and Datta, Gaurav and Chen, Lawrence Yunliang and Panitch, William Chung-Ho and Liu, Fangchen and Li, Hui and Goldberg, Ken},
  journal={2025 IEEE international conference on robotics and automation (ICRA)},
  year={2025}
}

@article{barreiros2025careful,
  title={A careful examination of large behavior models for multitask dexterous manipulation},
  author={{TRI LBM Team} and Jose Barreiros and Andrew Beaulieu and Aditya Bhat and Rick Cory and Eric Cousineau and Hongkai Dai and Ching-Hsin Fang and Kunimatsu Hashimoto and Muhammad Zubair Irshad and Masha Itkina and Naveen Kuppuswamy and Kuan-Hui Lee and Katherine Liu and Dale McConachie and Ian McMahon and Haruki Nishimura and Calder Phillips-Grafflin and Charles Richter and Paarth Shah and Krishnan Srinivasan and Blake Wulfe and Chen Xu and Mengchao Zhang and Alex Alspach and Maya Angeles and Kushal Arora and Vitor Campagnolo Guizilini and Alejandro Castro and Dian Chen and Ting-Sheng Chu and Sam Creasey and Sean Curtis and Richard Denitto and Emma Dixon and Eric Dusel and Matthew Ferreira and Aimee Goncalves and Grant Gould and Damrong Guoy and Swati Gupta and Xuchen Han and Kyle Hatch and Brendan Hathaway and Allison Henry and Hillel Hochsztein and Phoebe Horgan and Shun Iwase and Donovon Jackson and Siddharth Karamcheti and Sedrick Keh and Joseph Masterjohn and Jean Mercat and Patrick Miller and Paul Mitiguy and Tony Nguyen and Jeremy Nimmer and Yuki Noguchi and Reko Ong and Aykut Onol and Owen Pfannenstiehl and Richard Poyner and Leticia Priebe Mendes Rocha and Gordon Richardson and Christopher Rodriguez and Derick Seale and Michael Sherman and Mariah Smith-Jones and David Tago and Pavel Tokmakov and Matthew Tran and Basile Van Hoorick and Igor Vasiljevic and Sergey Zakharov and Mark Zolotas and Rares Ambrus and Kerri Fetzer-Borelli and Benjamin Burchfiel and Hadas Kress-Gazit and Siyuan Feng and Stacie Ford and Russ Tedrake},
  journal={Science Robotics},
  volume={11},
  number={113},
  pages={eaea6201},
  year={2026},
  publisher={American Association for the Advancement of Science}
}

@article{liu2022robot,
  title={Robot learning on the job: Human-in-the-loop autonomy and learning during deployment},
  author={Liu, Huihan and Nasiriany, Soroush and Zhang, Lance and Bao, Zhiyao and Zhu, Yuke},
  journal={The International Journal of Robotics Research},
  pages={02783649241273901},
  year={2022},
  publisher={SAGE Publications Sage UK: London, England}
}

@article{luo2025precise,
  title={Precise and dexterous robotic manipulation via human-in-the-loop reinforcement learning},
  author={Luo, Jianlan and Xu, Charles and Wu, Jeffrey and Levine, Sergey},
  journal={Science Robotics},
  volume={10},
  number={105},
  pages={eads5033},
  year={2025},
  publisher={American Association for the Advancement of Science}
}

@article{nachum2018data,
  title={Data-efficient hierarchical reinforcement learning},
  author={Nachum, Ofir and Gu, Shixiang Shane and Lee, Honglak and Levine, Sergey},
  journal={Advances in neural information processing systems},
  volume={31},
  year={2018}
}

@article{kitano2004biological,
  title={Biological robustness},
  author={Kitano, Hiroaki},
  journal={Nature Reviews Genetics},
  volume={5},
  number={11},
  pages={826--837},
  year={2004},
  publisher={Nature Publishing Group UK London}
}

@article{roa2015grasp,
  title={Grasp quality measures: review and performance},
  author={Roa, M{\'a}ximo A and Su{\'a}rez, Ra{\'u}l},
  journal={Autonomous robots},
  volume={38},
  number={1},
  pages={65--88},
  year={2015},
  publisher={Springer}
}

@inproceedings{ferrari1992planning,
  title={Planning optimal grasps},
  author={Ferrari, Carlo and Canny, John},
  booktitle={Proceedings., 1992 IEEE International Conference on Robotics and Automation, 1992.},
  volume={3},
  pages={2290--2295},
  year={1992},
  organization={IEEE}
}

@article{bohg2013datas,
  title={Data-driven grasp synthesis—a survey},
  author={Bohg, Jeannette and Morales, Antonio and Asfour, Tamim and Kragic, Danica},
  journal={IEEE Transactions on robotics},
  volume={30},
  number={2},
  pages={289--309},
  year={2013},
  publisher={IEEE}
}

@article{mahler2017dex,
  title={Dex-Net 2.0: Deep Learning to Plan Robust Grasps with Synthetic Point Clouds and Analytic Grasp Metrics},
  author={Mahler, Jeffrey and Liang, Jacky and Niyaz, Sherdil and Laskey, Michael and Doan, Richard and Liu, Xinyu and Aparicio, Juan and Goldberg, Ken},
  journal={Robotics: Science and Systems XIII},
  year={2017},
  publisher={Robotics: Science and Systems Foundation}
}

@article{hasson2012energy,
  title={Energy margins in dynamic object manipulation},
  author={Hasson, Christopher J and Shen, Tian and Sternad, Dagmar},
  journal={Journal of Neurophysiology},
  volume={108},
  number={5},
  pages={1349--1365},
  year={2012},
  publisher={American Physiological Society Bethesda, MD}
}

@article{sternad2016predictability,
  title={Predictability and robustness in the manipulation of dynamically complex objects},
  author={Sternad, Dagmar and Hasson, Christopher J},
  journal={Progress in Motor Control: Theories and Translations},
  pages={55--77},
  year={2016},
  publisher={Springer}
}

@article{dong2024characterizing,
  title={Characterizing manipulation robustness through energy margin and caging analysis},
  author={Dong, Yifei and Cheng, Xianyi and Pokorny, Florian T},
  journal={IEEE Robotics and Automation Letters},
  volume={9},
  number={9},
  pages={7525--7532},
  year={2024},
  publisher={IEEE}
}

@article{mahler2018synthesis,
  title={Synthesis of energy-bounded planar caging grasps using persistent homology},
  author={Mahler, Jeffrey and Pokorny, Florian T and Niyaz, Sherdil and Goldberg, Ken},
  journal={IEEE Transactions on Automation Science and Engineering},
  volume={15},
  number={3},
  pages={908--918},
  year={2018},
  publisher={IEEE}
}

@article{bazzi2020robustness,
  title={Robustness in human manipulation of dynamically complex objects through control contraction metrics},
  author={Bazzi, Salah and Sternad, Dagmar},
  journal={IEEE robotics and automation letters},
  volume={5},
  number={2},
  pages={2578--2585},
  year={2020},
  publisher={IEEE}
}

@inproceedings{maler2004monitoring,
  title={Monitoring temporal properties of continuous signals},
  author={Maler, Oded and Nickovic, Dejan},
  booktitle={International symposium on formal techniques in real-time and fault-tolerant systems},
  pages={152--166},
  year={2004},
  organization={Springer}
}

@inproceedings{donze2010robust,
  title={Robust satisfaction of temporal logic over real-valued signals},
  author={Donz{\'e}, Alexandre and Maler, Oded},
  booktitle={International conference on formal modeling and analysis of timed systems},
  pages={92--106},
  year={2010},
  organization={Springer}
}

@inproceedings{li2017reinforcement,
  title={Reinforcement learning with temporal logic rewards},
  author={Li, Xiao and Vasile, Cristian-Ioan and Belta, Calin},
  booktitle={2017 IEEE/RSJ International Conference on Intelligent Robots and Systems (IROS)},
  pages={3834--3839},
  year={2017},
  organization={IEEE}
}

@article{kapoor2020model,
  title={Model-based reinforcement learning from signal temporal logic specifications},
  author={Kapoor, Parv and Balakrishnan, Anand and Deshmukh, Jyotirmoy V},
  journal={arXiv preprint arXiv:2011.04950},
  year={2020}
}

@article{meng2023signal,
  title={Signal temporal logic neural predictive control},
  author={Meng, Yue and Fan, Chuchu},
  journal={IEEE Robotics and Automation Letters},
  volume={8},
  number={11},
  pages={7719--7726},
  year={2023},
  publisher={IEEE}
}

@inproceedings{takano2021continuous,
  title={Continuous optimization-based task and motion planning with signal temporal logic specifications for sequential manipulation},
  author={Takano, Rin and Oyama, Hiroyuki and Yamakita, Masaki},
  booktitle={2021 IEEE international conference on robotics and automation (ICRA)},
  pages={8409--8415},
  year={2021},
  organization={IEEE}
}

@inproceedings{varnai2020robustness,
  title={On robustness metrics for learning STL tasks},
  author={Varnai, Peter and Dimarogonas, Dimos V},
  booktitle={2020 American Control Conference (ACC)},
  pages={5394--5399},
  year={2020},
  organization={IEEE}
}

@article{dhonthi2021study,
  title={Study of signal temporal logic robustness metrics for robotic tasks optimization},
  author={Dhonthi, Akshay and Schillinger, Philipp and Rozo, Leonel and Nardi, Daniele},
  journal={arXiv preprint arXiv:2110.00339},
  year={2021}
}

@article{aljalbout2025reality,
  title={The reality gap in robotics: Challenges, solutions, and best practices},
  author={Aljalbout, Elie and Xing, Jiaxu and Romero, Angel and Akinola, Iretiayo and Garrett, Caelan Reed and Heiden, Eric and Gupta, Abhishek and Hermans, Tucker and Narang, Yashraj and Fox, Dieter and Scaramuzza, Davide and Ramos, Fabio},
  journal={Annual Review of Control, Robotics, and Autonomous Systems},
  volume={9},
  year={2025},
  publisher={Annual Reviews}
}

@article{gao2026taxonomy,
  title={A Taxonomy for Evaluating Generalist Robot Manipulation Policies},
  author={Gao, Jensen and Belkhale, Suneel and Dasari, Sudeep and Balakrishna, Ashwin and Shah, Dhruv and Sadigh, Dorsa},
  journal={IEEE Robotics and Automation Letters},
  year={2026},
  publisher={IEEE}
}

@article{firoozi2025foundation,
  title={Foundation models in robotics: Applications, challenges, and the future},
  author={Firoozi, Roya and Tucker, Johnathan and Tian, Stephen and Majumdar, Anirudha and Sun, Jiankai and Liu, Weiyu and Zhu, Yuke and Song, Shuran and Kapoor, Ashish and Hausman, Karol and Ichter, Brian and Driess, Danny and Wu, Jiajun and Lu, Cewu and Schwager, Mac},
  journal={The International Journal of Robotics Research},
  volume={44},
  number={5},
  pages={701--739},
  year={2025},
  publisher={SAGE Publications Sage UK: London, England}
}

@article{hunt2004crafting,
  title={The crafting of hook tools by wild New Caledonian crows},
  author={Hunt, Gavin R and Gray, Russell D},
  journal={Proceedings of the Royal Society of London. Series B: Biological Sciences},
  volume={271},
  number={suppl\_3},
  pages={S88--S90},
  year={2004},
  publisher={The Royal Society}
}

@article{gelblum2015ant,
  title={Ant groups optimally amplify the effect of transiently informed individuals},
  author={Gelblum, Aviram and Pinkoviezky, Itai and Fonio, Ehud and Ghosh, Abhijit and Gov, Nir and Feinerman, Ofer},
  journal={Nature communications},
  volume={6},
  number={1},
  pages={7729},
  year={2015},
  publisher={Nature Publishing Group UK London}
}

@article{catania1996unusual,
  title={The unusual nose and brain of the star-nosed mole},
  author={Catania, Kenneth C and Kaas, Jon H},
  journal={Bioscience},
  volume={46},
  number={8},
  pages={578--586},
  year={1996},
  publisher={JSTOR}
}

@article{shi2024rethinking,
  title={Rethinking Robustness Assessment: Adversarial Attacks on Learning-based Quadrupedal Locomotion Controllers},
  author={Shi, Fan and Zhang, Chong and Miki, Takahiro and Lee, Joonho and Hutter, Marco and Coros, Stelian},
  journal={Robotics: Science and System XX},
  year={2024},
  publisher={Robotics Science \& Systems Foundation}
}

@article{johnson2016hybrid,
  title={A hybrid systems model for simple manipulation and self-manipulation systems},
  author={Johnson, Aaron M and Burden, Samuel A and Koditschek, Daniel E},
  journal={The International Journal of Robotics Research},
  volume={35},
  number={11},
  pages={1354--1392},
  year={2016},
  publisher={SAGE Publications Sage UK: London, England}
}

@article{amato2025data,
  title={“Data will solve robotics and automation: True or false?”: A debate},
  author={Amato, Nancy M and Hutchinson, Seth and Garg, Animesh and Billard, Aude and Rus, Daniela and Tedrake, Russ and Park, Frank and Goldberg, Ken},
  journal={Science Robotics},
  volume={10},
  number={105},
  pages={eaea7897},
  year={2025},
  publisher={American Association for the Advancement of Science}
}

@article{to2001jamming,
  title={Jamming of granular flow in a two-dimensional hopper},
  author={To, Kiwing and Lai, Pik-Yin and Pak, HK},
  journal={Physical review letters},
  volume={86},
  number={1},
  pages={71},
  year={2001},
  publisher={APS}
}

@article{zhu2020robosuite,
  title={robosuite: A modular simulation framework and benchmark for robot learning},
  author={Zhu, Yuke and Wong, Josiah and Mandlekar, Ajay and Mart{\'\i}n-Mart{\'\i}n, Roberto and Joshi, Abhishek and Nasiriany, Soroush and Zhu, Yifeng},
  journal={arXiv preprint arXiv:2009.12293},
  year={2020}
}

@article{calli2017yale,
  title={Yale-CMU-Berkeley dataset for robotic manipulation research},
  author={Calli, Berk and Singh, Arjun and Bruce, James and Walsman, Aaron and Konolige, Kurt and Srinivasa, Siddhartha and Abbeel, Pieter and Dollar, Aaron M},
  journal={The International Journal of Robotics Research},
  volume={36},
  number={3},
  pages={261--268},
  year={2017},
  publisher={SAGE Publications Sage UK: London, England}
}

@article{correll2016analysis,
  title={Analysis and observations from the first amazon picking challenge},
  author={Correll, Nikolaus and Bekris, Kostas E and Berenson, Dmitry and Brock, Oliver and Causo, Albert and Hauser, Kris and Okada, Kei and Rodriguez, Alberto and Romano, Joseph M and Wurman, Peter R},
  journal={IEEE Transactions on Automation Science and Engineering},
  volume={15},
  number={1},
  pages={172--188},
  year={2016},
  publisher={IEEE}
}

@inproceedings{zahid2024cloudgripper,
  title={Cloudgripper: An open source cloud robotics testbed for robotic manipulation research, benchmarking and data collection at scale},
  author={Zahid, Muhammad and Pokorny, Florian T},
  booktitle={2024 IEEE International Conference on Robotics and Automation (ICRA)},
  pages={12076--12082},
  year={2024},
  organization={IEEE}
}

@article{chen2026manipulationnet,
  title={ManipulationNet: An Infrastructure for Benchmarking Real-World Robot Manipulation with Physical Skill Challenges and Embodied Multimodal Reasoning},
  author={Chen, Yiting and Kimble, Kenneth and Adelson, Edward H. and Asfour, Tamim and Chanrungmaneekul, Podshara and Chitta, Sachin and Chitambar, Yash and Chen, Ziyang and Goldberg, Ken and Kragic, Danica and Li, Hui and Li, Xiang and Li, Yunzhu and Prather, Aaron and Pollard, Nancy and Roa-Garzon, Maximo A. and Seney, Robert and Sha, Shuo and Wang, Shihefeng and Xiang, Yu and Zhang, Kaifeng and Zhu, Yuke and Hang, Kaiyu},
  journal={arXiv preprint arXiv:2603.04363},
  year={2026}
}

@article{kroemer2021review,
  title={A review of robot learning for manipulation: Challenges, representations, and algorithms},
  author={Kroemer, Oliver and Niekum, Scott and Konidaris, George},
  journal={Journal of machine learning research},
  volume={22},
  number={30},
  pages={1--82},
  year={2021}
}

@inproceedings{florence2018dense,
  title={Dense Object Nets: Learning Dense Visual Object Descriptors By and For Robotic Manipulation},
  author={Florence, Peter R and Manuelli, Lucas and Tedrake, Russ},
  booktitle={Conference on Robot Learning},
  pages={373--385},
  year={2018},
  organization={PMLR}
}

@inproceedings{xiong2025vision,
  title={Vision in Action: Learning Active Perception from Human Demonstrations},
  author={Xiong, Haoyu and Xu, Xiaomeng and Wu, Jimmy and Hou, Yifan and Bohg, Jeannette and Song, Shuran},
  booktitle={Conference on Robot Learning},
  pages={5450--5463},
  year={2025},
  organization={PMLR}
}

@misc{fox_grasping_slides,
  author       = {Fox, Dieter},
  title        = {Grasping},
  howpublished = {Lecture slides, CSE 478: Robot Learning, University of Washington},
  year         = {2020},
  url          = {https://courses.cs.washington.edu/courses/cse478/20wi/site/resources/lec23_grasping.pdf}
}

@article{rodriguez2012caging,
  title={From caging to grasping},
  author={Rodriguez, Alberto and Mason, Matthew T and Ferry, Steve},
  journal={The International Journal of Robotics Research},
  volume={31},
  number={7},
  pages={886--900},
  year={2012},
  publisher={SAGE Publications Sage UK: London, England}
}

@article{song2017controllable,
  title={Controllable load sharing for soft adhesive interfaces on three-dimensional surfaces},
  author={Song, Sukho and Drotlef, Dirk-Michael and Majidi, Carmel and Sitti, Metin},
  journal={Proceedings of the National Academy of Sciences},
  volume={114},
  number={22},
  pages={E4344--E4353},
  year={2017},
  publisher={National Academy of Sciences}
}

@article{kress2024robot,
  title={Robot learning as an empirical science: Best practices for policy evaluation},
  author={Kress-Gazit, Hadas and Hashimoto, Kunimatsu and Kuppuswamy, Naveen and Shah, Paarth and Horgan, Phoebe and Richardson, Gordon and Feng, Siyuan and Burchfiel, Benjamin},
  journal={arXiv preprint arXiv:2409.09491},
  year={2024}
}

@article{liconti2026benchmark,
  title={A Benchmark of Dexterity for Anthropomorphic Robotic Hands},
  author={Liconti, Davide and Zhou, Yuning and Toshimitsu, Yasunori and Hinchet, Ronan and Katzschmann, Robert K},
  journal={arXiv preprint arXiv:2604.09294},
  year={2026}
}

@article{kang2026learning,
  title={Learning Force-Regulated Manipulation with a Low-Cost Tactile-Force-Controlled Gripper},
  author={Kang, Xuhui and Tian, Tongxuan and Lee, Sung-Wook and Huang, Binghao and Li, Yunzhu and Kuo, Yen-Ling},
  journal={arXiv preprint arXiv:2602.10013},
  year={2026}
}

@inproceedings{atreya2025roboarena,
  title={RoboArena: Distributed Real-World Evaluation of Generalist Robot Policies},
  author={Atreya, Pranav and Pertsch, Karl and Lee, Tony and Kim, Moo Jin and Jain, Arhan and Kuramshin, Artur and Neary, Cyrus and Hu, Edward S and Arora, Kanav and Ellis, Kirsty and Macesanu, Luca and Leonard, Matthew and Cho, Meedeum and Aslan, Ozgur and Dass, Shivin and Wang, Tony and Yuan, Xingfang and Gupta, Abhishek and Jayaraman, Dinesh and Berseth, Glen and Daniilidis, Kostas and Mart{'i}n-Mart{'i}n, Roberto and Lee, Youngwoon and Liang, Percy and Finn, Chelsea and Levine, Sergey},
  booktitle={Conference on Robot Learning},
  pages={336--364},
  year={2025},
  organization={PMLR}
}

@article{kasper2012kit,
  title={The kit object models database: An object model database for object recognition, localization and manipulation in service robotics},
  author={Kasper, Alexander and Xue, Zhixing and Dillmann, R{\"u}diger},
  journal={The International Journal of Robotics Research},
  volume={31},
  number={8},
  pages={927--934},
  year={2012},
  publisher={SAGE Publications Sage UK: London, England}
}

@article{tao2024maniskill3,
  title={Maniskill3: Gpu parallelized robotics simulation and rendering for generalizable embodied ai},
  author={Tao, Stone and Xiang, Fanbo and Shukla, Arth and Qin, Yuzhe and Hinrichsen, Xander and Yuan, Xiaodi and Bao, Chen and Lin, Xinsong and Liu, Yulin and Chan, Tse-kai and Gao, Yuan and Li, Xuanlin and Mu, Tongzhou and Xiao, Nan and Gurha, Arnav and Huang, Zhiao and Calandra, Roberto and Chen, Rui and Luo, Shan and Su, Hao},
  journal={arXiv preprint arXiv:2410.00425},
  year={2024}
}

@article{james2020rlbench,
  title={Rlbench: The robot learning benchmark \& learning environment},
  author={James, Stephen and Ma, Zicong and Arrojo, David Rovick and Davison, Andrew J},
  journal={IEEE Robotics and Automation Letters},
  volume={5},
  number={2},
  pages={3019--3026},
  year={2020},
  publisher={IEEE}
}

\appendix
% \section{Implementation Details}

\end{document}